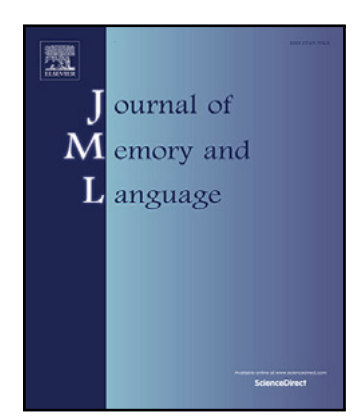

# LLMs as models for analogical reasoning

Sam Musker [a],*, Alex Duchnowski [a], Raphaël Millière [b], Ellie Pavlick [a]

[a] Brown University, Department of Computer Science, Providence, 02912, RI, USA

[b] Macquarie University, Department of Philosophy, Sydney, 2109, NSW, Australia

## ARTICLE INFO

## ABSTRACT

Analogical reasoning — the capacity to identify and map structural relationships between different domains — is fundamental to human cognition and learning. Recent studies have shown that large language models (LLMs) can sometimes match humans in analogical reasoning tasks, opening the possibility that analogical reasoning might emerge from domain-general processes. However, it is still debated whether these emergent capacities are largely superficial and limited to simple relations seen during training or whether they encompass the flexible representational and mapping capabilities which are the focus of leading cognitive models of analogy. In this study, we introduce novel analogical reasoning tasks that require participants to map between semantically contentful words and sequences of letters and other abstract characters. This task necessitates the ability to flexibly *re-represent* rich semantic information—an ability which is known to be central to human analogy but which is thus far not well-captured by existing cognitive theories and models. We assess the performance of both human participants and LLMs on tasks focusing on reasoning from semantic structure and semantic content, introducing variations that test the robustness of their analogical inferences. Advanced LLMs match human performance across several conditions, though humans and LLMs respond differently to certain task variations and semantic distractors. Our results thus provide new evidence that LLMs might offer a *how-possibly* explanation of human analogical reasoning in contexts that are not yet well modeled by existing theories, but that even today's best models are unlikely to yield *how-actually* explanations.

## Introduction

Analogical reasoning — the ability to identify and map structural relationships between a source and target domain — is posited to play a central role in human learning and generalization. It enables us to transfer knowledge from familiar domains to unfamiliar ones, recognize abstract patterns across superficially different contexts, and generate creative solutions to novel problems. Analogy is often credited with giving humans the ability to reason efficiently in unfamiliar domains (Hofstadter, 2001; Holyoak, Gentner, & Kokinov, 2001) and is considered by some to be among the primary predictors of intelligence and higher-order thinking (Gentner & Rattermann, 1994; Hesse, 1966). As a result, cognitive scientists have long sought to understand the computational mechanisms by which humans carry out this process.

With recent advances in artificial intelligence, there has been a growing interest in whether large language models (LLMs) possess a human-like capacity for analogical reasoning. Unlike existing computational models of analogy, LLMs contain no innate structures or symbolic mechanisms designed for analogical reasoning, and are not explicitly trained on analogical reasoning tasks. Rather, the behaviors that LLMs exhibit are the consequence of their relatively passive learning objectives, namely next-word prediction over large text corpora and, optionally, supervised fine-tuning for instruction-following and conversation. This raises the question of whether the capacity for analogical reasoning could emerge as a by-product of domain-general statistical learning, and if so, whether this could constrain theories of analogical reasoning in humans. Admittedly, the way LLMs are trained does not reflect the developmental conditions under which humans acquire analogical reasoning. They require orders of magnitude more language data than children are estimated to receive before achieving fluency (Gilkerson et al., 2017; Hu et al., 2024) and typically rely on a developmentally unrealistic mix of text and computer code scraped from the internet. Even so, their success at analogical reasoning has important implications for cognitive theory: if analogical reasoning can emerge as a by-product of self-supervised learning and a generic neural architecture, it raises the burden of proof for computational and theoretical work on analogy which posits richer innate structure or more complex learning signals.

* Corresponding author.
*E-mail addresses:* samuel_musker@brown.edu (S. Musker), alex_duchnowski@brown.edu (A. Duchnowski), raphael.milliere@mq.edu.au (R. Millière), ellie_pavlick@brown.edu (E. Pavlick).

Work to date has yielded mixed interpretations of whether LLMs do in fact achieve human-level analogical reasoning. In 2023, Webb et al. found that GPT-3 could match or exceed typical human performance on most analogical tasks, including verbal analogies (e.g., 'drive : car :: burn : ?'), letter-string analogies (e.g., transforming 'a b c d' to 'a b c e' and applying similar patterns to other sequences), and digit matrix problems (e.g., completing 3 × 3 matrices of digits following specific patterns or rules) (Webb, Holyoak, & Lu, 2023). Additionally, they found that GPT-4 showed strong performance on story analogies requiring the model to identify similar patterns between different stories, even when the stories described very different situations. Across the range of analogical reasoning problems tested, Hu, Storks, Lewis, and Chai (2023) and Webb et al. (2023) find that LLMs, in particular the most recent and advanced models, not only match human performance but replicate human-like error patterns as well.

However, contemporaneous work argues that LLMs' abilities are overstated, and that their apparent analogical reasoning abilities might have alternative explanations. For example, LLMs' performance degrades substantially when they are tested on stimuli that are deliberately designed to ensure that they are unlikely to appear in training data, whereas human performance remains stable (Lewis & Mitchell, 2024). It has yet to be determined whether this discrepancy is actually diagnostic of a failure of LLMs to engage in genuine analogical reasoning or whether it can be attributed instead to auxiliary task demands (Webb, Holyoak, & Lu, 2024). It thus remains an open question to what extent LLMs should be viewed as human-level or human-like with regard to analogy, and if so, whether LLMs' success can advance cognitive theory in a meaningful way. The present state of this debate thus calls for novel analogical reasoning tasks that are likely not present in LLMs' training data and which focus on aspects that have proven challenging for existing theoretical frameworks.

One such challenging aspect is flexible *re-representation*, the ability to dynamically restructure how concepts are encoded to support mapping between domains (Chalmers, French, & Hofstadter, 1992). For example, a semantic content analogy from our stimuli is as follows:

```
horse => * * * *
cat => * * * *
ant => ! ! ! ! ! !
bee => ! ! ! ! ! !
chicken => ! !
spider => ! ! ! ! ! ! ! !
dog => * * * *
human =>
```

Completing this analogy requires inferring the abstract rule in which concepts on the left-hand side are represented in terms of two features: NUMBER-OF-LEGS and IS-MAMMAL, with these features respectively determining the number and type of symbol which appears on the right-hand side. Since the concepts on the left-hand side have many features and it is not known at the outset which ones will be relevant to the mapping, a system for solving the task must possess the ability to dynamically adjust which features are used to represent the concepts. At the local level, individual concepts must often be re-represented in different ways depending on the analogy being drawn. For instance, the concept of 'dog' might need to be encoded in terms of its taxonomic category, physical attributes, diet, or domestication status. At a more global level, solving such tasks might invoke processes for schema induction (Rumelhart & Norman, 1976) in order to recognize that superficially different scenarios instantiate the same abstract relational category or pattern.

While there exist a number of compelling cognitive theories of analogy, none readily account for the type of reasoning required by the above task, either because they do not address re-representation, or because they cannot be applied to real-world semantic concepts. For example, Structure Mapping Theory (SMT) emphasizes the importance of systematic relational mapping between structured representations (Gentner, 1983). Although successful in explaining many aspects of human analogical reasoning, particularly the preference for mapping systems of relations rather than isolated features, SMT largely sidesteps the question of how appropriate representations are constructed. The theory has been criticized for assuming that concepts are already represented in predicate-logic form with clearly specified relations and thus failing to explain how humans identify which relations are relevant for a given analogy in the first place (Chalmers et al., 1992).

The first model which attempted to address the question of re-representation is Copycat (Hofstadter & Mitchell, 1995). Copycat was directly designed to account for the need to dynamically explore the space of possible representations in the course of analogy making and accomplished this by combining bottom-up and top-down processes. This facilitated re-representing the same letter string in multiple ways depending on the patterns being mapped. However, it has only been successfully applied to synthetic domains such as letter-string analogies. While the approach is intended to be domain-general in principle, in practice it would require significant theoretical and engineering effort to be adapted to the open-domain concepts we employ here, if it is in fact possible to do so.

More recent approaches have attempted to bridge this gap. The BART (Bayesian Analogy with Relational Transformations) model accounts for the emergence of relational representations from non-relational inputs through statistical learning (Lu, Wu, & Holyoak, 2019). Other work has made progress on modeling re-representation in specific domains, such as visual patterns in Raven's Progressive Matrices (Lovett & Forbus, 2017) or abstract concepts through program induction (Rule, Schulz, Piantadosi, & Tenenbaum, 2018). These models, like Copycat, include a set of bottom-up processes which can respond to top-down signals in order to account for re-representation in the course of analogy formation. Once again, these approaches remain limited to relatively constrained domains and have not been extended to handle open-ended semantic concepts.

This points to an important gap in our understanding of analogical inference: no current framework adequately explains how humans flexibly re-represent semantically rich concepts and relations to support novel analogical mappings and extract generalizable rules. There is thus a need for a computational model which can explain human behavior on complex analogical reasoning tasks using domain-general mechanisms.

To test whether LLMs might address this gap, we design a novel set of analogical reasoning tasks (split into two studies) which we present to both humans and LLMs. Our tasks focus respectively on reasoning from *semantic structure* and *semantic content*. Study 1 examines how subjects leverage semantic structure (relationships between concepts) to reason about arbitrary symbolic patterns in a target domain. Through systematic manipulation of available information — including conditions that remove linguistic content entirely or introduce misleading semantic relationships — we can isolate the role that semantic structure plays in guiding analogical inference. Study 2 extends this investigation by focusing on semantic content rather than just structural relationships. Subjects must identify relevant features of concepts (like NUMBER-OF-LEGS and IS-MAMMAL) and map these onto abstract relations between arbitrary symbols. We manipulate the nature and number of attributes required to perform the analogical inference to assess how this affects performance in humans and LLMs.

Our experiments show that LLMs might provide a how-possibly model of non-trivial analogical reasoning in human subjects: the most successful LLMs achieve human-level performance on our challenging tasks which require abstract rule induction and re-representation. However, we also find that, while LLMs can reach human-level performance, even the best models demonstrate significantly different patterns of behavior in response to control conditions and stress tests, making them unlikely to yield how-actually explanations.[1] In Study 1, we find that LLMs are more sensitive to presentation order and struggle when irrelevant semantic distractors are present. In Study 2, the best performing LLM matched human performance across all conditions, and exceeded it in the compositional variants on which human subjects performed most poorly.

Taken together, our findings suggest that LLMs can achieve sophisticated analogical reasoning capabilities through domain-general learning mechanisms, although they likely employ different mechanisms than humans. This suggests that LLMs are poised to offer insights on behaviors that current cognitive models cannot explain, and that research seeking to unpack the more precise mechanisms LLMs employ in order to achieve this behavior, while challenging, is likely to be a fruitful endeavor for cognitive theory.

## Analogical representation tasks

### *Study 1: Analogical inference from Semantic Structure*

Study 1 investigates the capacity to map semantic relationships between concepts in a source domain to symbolic patterns in a non-semantic target domain. The task design specifically relates to the *inference process* in analogical reasoning, rather than to the initial mapping discovery that characterizes many traditional analogy tasks, and has a particular focus on how semantic structure can guide analogical inference in formal domains.

The core mechanism underlying the tasks in Study 1 involves a structural mapping where subjects must: (1) identify meaningful relations in the source domain, (2) abstract these relations into a form that can be mapped onto the target domain, and (3) generate appropriate completions by applying the mapped relations. Subjects are presented with several rows showing explicit mappings between words (source domain) and symbol sequences (target domain), followed by a final row containing only a word. Their task is to infer the corresponding symbol sequence that completes the pattern.

In the Defaults condition, subjects might see “dog” mapped to “X”, “puppy” mapped to “x”, and “cat” mapped to “Y”, and be asked to produce the mapping for “kitten”. To succeed at this task, subjects must first re-represent the concepts to identify relevant relational patterns (e.g., recognizing that both dog ↔ puppy and cat ↔ kitten instantiate an adult ↔ juvenile relationship). Second, they must induce a general rule that maps this semantic relationship onto the formal pattern in the symbol sequences (e.g., adult → uppercase, juvenile → lowercase). A question from this condition is illustrated in Fig. 1 along with the instructions shown to participants.

We then include several control conditions. The first set of these controls is intended to probe the reasoning mechanism, in particular testing how it responds to perturbations and distractions in how the questions are presented. Prior work suggests such changes could influence performance by making the relevant relations less transparent: see, for example, work on the blocking advantage in humans (Carvalho & Goldstone, 2012) and in LLMs (Russin, Pavlick, & Frank, 2024).

The Permuted Pairs condition presents the same mappings in scrambled order, testing whether subjects can extract the underlying rule regardless of presentation sequence. This condition tests whether subjects' performance is robust to the order in which example mappings are provided. While the Permuted Pairs condition changes the order of lines within each question to disrupt the analogy structure, the Permuted Questions condition leaves questions completely intact but shuffles the order in which the questions are presented in order to verify that the arbitrary question ordering does not have an unintended effect.

**Table 1**
Defaults and control conditions used to measure ability of humans and LLMs to perform analogical reasoning tasks that involve analogical reasoning from a linguistic to a non-linguistic domain. The Permuted Questions condition (not shown) is identical to Defaults, but with question order permuted.

| | | |
|---|---|---|
| Defaults | Basic test of analogical reasoning from a linguistic to a non-linguistic domain | square => C C C<br>rectangle => c c c<br>circle => C C<br>oval => |
| Permuted Pairs | Like Defaults, but with row order permuted | rectangle => c c c<br>circle => C C<br>square => C C C<br>oval => |
| Distracted | Like Defaults, but with a distractor row added | square => C C C<br>rectangle => c c c<br>pillow => A P<br>circle => C C<br>oval => |

**Table 2**
Conditions involving alteration or omission of the source domain. The Random Permuted Pairs condition (not shown) is identical to Randoms, but with the order of elements within questions permuted.

| | | |
|---|---|---|
| Only RHS | Test of how well the answer can be inferred without using the source domain | C C C<br>c c c<br>C C |
| Randoms | Variant of Defaults in which there is no semantic structure relating the words on the left-hand side | banana => C C C<br>fireplace => c c c<br>bean => C C<br>plug => |
| Random Finals | Variant of Defaults in which the final term is not semantically related to the preceding terms | square => C C C<br>rectangle => c c c<br>circle => C C<br>lime => |

The Distracted condition introduces irrelevant mappings, requiring subjects to identify and suppress information that does not fit the core analogical pattern. This condition tests whether subjects are able to ignore irrelevant mappings and rely exclusively on relevant ones for inference. The Defaults, Permuted Pairs, and Distracted conditions are shown in Table 1.

Our second set of controls is designed to diagnose the extent to which subjects actually leverage semantic structure in the source domain to solve these tasks.

The Only RHS condition removes the words entirely, presenting only the symbol sequences. This condition tests whether subjects can complete the patterns through purely formal reasoning without semantic guidance.

The Randoms condition uses semantically unrelated words in the source domain, testing whether subjects will identify the semantic content on the left-hand side as “misleading” and thus focus exclusively on symbolic patterns in the target domain.

By contrast, the Random Finals condition initially establishes a semantic pattern but breaks it in the final term, testing whether subjects will dynamically shift their strategy when semantic structure appears to be “misleading”.

We interpret a performance discrepancy between either the Random or Random Finals conditions and the Only RHS condition as evidence that the subject has a bias toward using semantic structure even when ignoring it altogether would enable a simpler representation of the pattern. See Table 2 for examples from these conditions.

The $2 \times n$ condition provides the strongest test of schema induction by requiring the integration of multiple relationships to determine the correct mapping (see the example in Table 3). For instance, subjects

[1] In philosophy of science, a how-possibly explanation outlines a plausible mechanism that could generate the phenomenon, whereas a how-actually explanation specifies the mechanism that does in fact generate it (Brandon, 1990).

We are conducting an experiment on general reasoning abilities. Below we will show you various words and drawings of each, after which you will need to complete the last drawing. Respond as concisely as possible with only the last drawing.
Question 1:
square => C C C
rectangle => c c c
circle => C C
oval =>

**Fig. 1.** An example question from the Defaults condition of the Semantic Structure experiment. A subject observes four questions in a row from one condition only with the answers to prior questions revealed (we call this sequence of four questions a "quiz").

**Table 3**
The $2 \times n$ condition, used to diagnose subjects' tendency to rely on RHS-only heuristics to solve the task. In this example, reasonable answers based on only the right-hand side include H, X, Z#Z, and M#M to complete pairs in accordance with the relationship between V and V#V. These four responses give a baseline RHS-only success of 25%. All questions in this condition share this feature of having a low baseline probability of success if only the right-hand side is used.

| | | |
|---|---|---|
| $2 \times n$ | A condition in which words from a variety of consistent pairs are shown, with a single complete pair ('torso' and 'shirt' in the example) revealing how the relation is encoded on the RHS. | pants => H # H<br>glove => X # X<br>torso => V<br>foot => Z<br>head => M<br>shirt => V # V<br>hat => |

might need to understand both what body parts are covered by what clothing items in the source domain and how this relationship determines symbol patterns in the target domain. This condition prevents a solution through simple pattern completion or one-to-one mapping, instead requiring subjects to induce and apply a higher-order relational rule.

### Study 2: Analogical inference from Semantic Content

While Study 1 focused on mapping structural relations, Study 2 investigates how subjects identify relevant semantic properties of individual concepts and induce rules that systematically connect these properties to formal patterns. Importantly, none of the tasks of Study 2 can be solved through pure pattern matching by considering only symbol sequences in the target domain.

The study includes four conditions that vary the type and number of semantic properties governing the mappings. Each condition requires subjects to identify relevant semantic properties and use them to guide inference, rather than applying transformation rules as in Study 1.[2]

In the Categorial condition, subjects must identify a single categorial distinction that determines which symbol is used in the mapping. For example, participants might see various animals mapped to either '∗' or '!', where the pattern depends on whether each animal is a mammal. Success requires searching through possible categorial distinctions (such as vertebrate/invertebrate, flying/non-flying, or mammal/non-mammal) to identify which one accounts for the observed symbol assignments.

The Dual-attribute condition increases complexity by requiring participants to track two independent categorial properties simultaneously. These properties jointly determine both the type of symbol(s) used in the target domain and the pattern they form. For example, subjects might see family members mapped to sequences of symbols where gender determines the symbol type (male → '!', female → '∗') and generation number determines the sequence length (grandparent → 1, parent → 2, child → 3). This condition tests participants' ability to decompose complex patterns into independent components and identify relevant semantic features for each component.

The Numeric condition focuses specifically on quantitative semantic properties, requiring participants to identify numeric features that determine the length of symbol sequences. For example, participants might see various animals mapped to sequences of asterisks where the sequence length matches the number of legs each animal has. This condition tests subjects' abilities to re-represent concepts in the source domain in terms of their quantitative properties and identify which specific quantitative property explains the observed formal patterns in the target domain.

Finally, the Numeric Dual-attribute condition combines categorial and quantitative reasoning. Subjects must simultaneously identify a categorial feature that determines symbol choice and a numeric feature that determines sequence length. Like the Dual-attribute condition, this condition is particularly demanding as it requires inducing and integrating two independent rules while filtering out irrelevant properties of the concepts. These conditions are shown in Table 4.

Across all conditions, the central challenge is again that of flexible re-representation: participants must dynamically adjust how they represent familiar concepts in the source domain based on the patterns they observe in the target domain. This requires suppressing typical ways of thinking about concepts in favor of task-relevant features. For example, when thinking about animals, one might typically focus on habitat, diet, or behavior, but the task might require representing them solely in terms of anatomical features like number of legs or reproductive characteristics.

## Methods

### Experiment details

In the Semantic Structure experiment, each subject was presented with a quiz, which is a sequence of four such questions generated using four sets of base domains and four sets of target domains selected such that a participant sees each base and target domain exactly once. Eight variants of the task were devised to investigate the influence of task variations as described above.

Questions are introduced with the prompt "We are conducting an experiment on general reasoning abilities. Below we will show you various words and drawings of each, after which you will need to complete the last drawing. Respond as concisely as possible with only the last drawing". Here, the term "drawings" directs the subject to complete the right-hand side on the basis of the left, without directly referencing "analogical reasoning". The experiment is introduced to human subjects and LLMs as studying "general reasoning abilities" for the same reason.

[2] While participants could use a property-based approach to solve the Semantic Structure problems, the Semantic Content problems are differentiated in that this is the only approach that will yield the reference answers.

**Table 4**
The conditions of the Semantic Content experiment.

| | | |
|---|---|---|
| Categorial | Right-hand terms are single characters corresponding to a Categorial property of the left-hand terms. | chicken => !<br>spider => !<br>cat => *<br>horse => *<br>ant => !<br>dog => *<br>bee => !<br>human => |
| Dual-attribute | Right-hand terms are a sequence of several characters that vary according to two properties of the left-hand terms. | grandfather => !<br>grandmother => *<br>mother => * *<br>father => ! !<br>brother => ! ! !<br>sister => |
| Numeric | Right-hand terms are a sequence of a single repeated character, with the number of repetitions corresponding to a numeric property of the left-hand terms. | chicken => * *<br>human => * *<br>dog => * * * *<br>spider => * * * * * *<br>* *<br>cat => * * * *<br>horse => * * * *<br>bee => |
| Numeric Dual-attribute | Right-hand terms are a sequence of a repeated character, with the number of repetitions corresponding to a numeric property of the left-hand terms and the character corresponding to a Categorial property. | horse => * * * *<br>cat => * * * *<br>ant => ! ! ! ! ! !<br>bee => ! ! ! ! ! !<br>chicken => ! !<br>spider => ! ! ! ! ! ! !<br>!<br>dog => * * * *<br>human => |

In the Semantic Content experiment, each condition (described in Table 4) contains two quizzes, with four questions per quiz. Unless otherwise stated, methodological details of the Semantic Content experiment match those of the Semantic Structure experiment.

*Participants*

*LLMs*

We run our experiments on the following LLMs: GPT-3 (`text-davinci-002`) (Brown et al., 2020), GPT-4 (`gpt-4-0314`) (OpenAI et al., 2024), Pythia-12B (Biderman et al., 2023), Claude 2 (Anthropic, 2023), Claude 3 Opus (Anthropic, 2024), Falcon-40B (Technology Innovation Institute, 2023), and Llama-405B (Grattafiori et al., 2024). All of the above are transformer-based LLMs trained primarily on a next-word prediction objective. OpenAI models are queried through the OpenAI API, Anthropic models are queried through the Anthropic API, Pythia is run on a local GPU server, and Llama is run through the Fireworks API. Across all models, we use a temperature of 0.7, and top_p is set to 1.

GPT-3 consists of a 175B parameter model trained on text completion and fine-tuned to produce more coherent answers. The details of GPT-4 are not publicly known, but it is considered by some sources to be a mixture-of-experts model consisting of numerous GPT-3-scale language models (Liu et al., 2023). GPT-4, unlike the GPT-3 Davinci-002 model used for testing, supplements text-completion pre-training and fine-tuning with reinforcement learning from human feedback (RLHF) in order to better align model outputs with the expectations of a human user. The Davinci-002 version of GPT-3 employs a "similar but slightly different" response alignment method to RLHF, also using human data (Janus, 2022; OpenAI, 2022). The training of Claude 2 includes RLHF, but its performance falls short of GPT-4. The more recent Claude 3 (in our case, the most advanced Opus version) is considered to approximately match GPT-4 performance in general. GPT-3 and -4 are developed by OpenAI, whereas Claude 2 and 3 are developed by Anthropic. Pythia-12B and Falcon-40B are open-weights LLMs trained on a text-completion objective and consist of 12B and 40B parameters respectively. Neither undergoes RLHF. Pythia-12B is developed by EleutherAI, and Falcon-40B is developed by the Technology Innovation Institute. The 405B parameter Llama 3 model, developed by Meta, is among the best available open-weights models. It is a relatively straightforward transformer architecture and is not a mixture of experts. It undergoes post-training based on supervised fine-tuning and other techniques.

*Human subjects*

We also test human participants on our experiments. Reported in the main text are results obtained from 194 (mostly undergraduate) Brown University students (132 in the Semantic Structure experiment, and 62 in the Semantic Content experiment). The split of participants between experiments approximately matches the 9:4 ratio of experiment conditions. The number of participants by condition are as follows: Defaults 18, Distracted 18, Only RHS 18, Permuted Pairs 17, Permuted Questions 17, Random Finals 15, Random Permuted Pairs 6, Randoms 8, $2 \times n$ 15, Categorial 16, Dual-attribute 16, Numeric 16, Numeric Dual-attribute 14. The Defaults, Distracted, Only RHS, Permuted Pairs, Permuted Questions, and Random Finals conditions each have four quizzes per condition while the remaining conditions each have two quizzes. Subjects were assigned randomly to a single quiz from one condition without the re-use of subjects. Roughly the same number of participants were assigned to each condition, with the exception of the Randoms and Random Permuted Pairs conditions. These were together assigned roughly the expected number of subjects for a single condition due to their similarity.

The subjects were recruited through email advertisements and offered $10 in compensation. Earlier results obtained for the Semantic Structure experiment from an online sample of participants recruited through Prolific are reported in Fig. A.11 of Appendix.

We ensure that humans and LLMs are given comparable information in our prompting design. A given human participant sees one quiz with four questions, with questions revealed one at a time with the answer shown following each response. LLMs are prompted with the first question of a quiz, then the second question with the first question and its (correct) answer accumulating in the prompt, and so forth for the four questions in a quiz. This prompt accumulation mimics the availability in the memory of human subjects of previous answers within a quiz.

*Statistical testing*

In each experiment, we are interested in the relative performance of human subjects and the best-performing models and how this depends on the particular experiment conditions. Differences between most models and human subjects are large and do not require statistical analysis, and so we focus our statistical analysis on the performance of GPT-4, Claude 3, and Llama-405B, each relative to human subjects.

For each experiment and pair of subjects (human subjects and GPT-4, human subjects and Claude 3, and human subjects and LLama-405B) we fit a logistic regression model to the data with and without interactions between the subject type and the experiment condition. In all cases, the outcome variable is the un-aggregated per-question score achieved by a subject (either a 0 or 1), and the predictor variables are experiment condition (e.g. "Defaults" or "Permuted Pairs") and subject type (e.g. "human subjects" or "GPT-4"). We use six likelihood ratio tests to assess whether the interaction between subject type and experiment condition is significant for a given pair of subjects within a particular experiment, as motivated by Glover (Glover & Dixon, 2004). In all cases, the interaction is significant, and so we use simple effects analysis to investigate the direction and significance of the effect of subject type within particular conditions.

For the Semantic Content experiment, we additionally perform a logistic simple effects analysis comparing the performance of a single subject type (human, GPT-4, Claude 3, or Llama-405B) in compositional versus non-compositional conditions for the numeric and non-numeric cases respectively with the non-compositional condition as reference. For example, we assess the effect Dual-attribute condition against the Categorial reference condition specifically when Claude 3 is the subject (and likewise for the other three examined subject types).

Standard errors are computed first per subject type per question number and then propagated to the condition level for each subject type. To find the standard error per question, we aggregate over the responses of different individuals; for models we aggregate over multiple samples from the same model.

Further details are provided in Appendices “Regression results, semantic structure experiment” and “Regression results, semantic content experiment” of Appendix.

## Results

### *Reasoning from Semantic Structure*

Recall (from Section “Study 1: Analogical inference from Semantic Structure”) that our Semantic Structure experiments are designed to assess analogical inference by asking subjects to map a set of words containing relational structure in the source domain to a set of strings related via non-linguistic string operations in the target domain. We describe both humans’ and LLMs’ overall performance first (Section “Overall performance”) before interpreting how performance changes within each of the more nuanced control conditions (Sections “Sensitivity to question presentation”–“Other observations”).

#### *Overall performance*

Human subjects perform well overall in the Defaults setting, with a proportion of between 0.4 and 0.9 of answers matching the reference across the various conditions.[3] The most advanced LLMs that we test match references in the range 0.1–0.95 in the different conditions we assess. This performance range is comparable to prior work on analogical reasoning over arbitrary symbols. For example, the results of human subjects on the “zero-generalization setting” studied by both Lewis and Mitchell (2024) and Webb et al. (2023) range from 0.2–0.8 in the former study and from 0.5–1.0 in the latter study. Similarly, results for LLMs (GPT-3, GPT-3.5, and GPT-4) across those conditions range from 0.1–1.0 in the two studies. Thus, our data suggest that our analogy problems are not inherently easier or harder than those used in prior work.

Fig. 2 shows the performance of humans and LLMs in the Defaults condition as a function of their performance on MMLU,[4] a widely-used language competency benchmark. Increasing MMLU score is associated with more matches to reference answers on the Defaults condition. This steadily increasing performance is presumed to correlate with the scale of model parameters and training data (Kaplan et al., 2020). Smaller models do not perform competitively (Pythia-12B obtains a reference match of 0.0, Falcon 40B 0.1, GPT-3 0.5, and Claude 2 0.6).

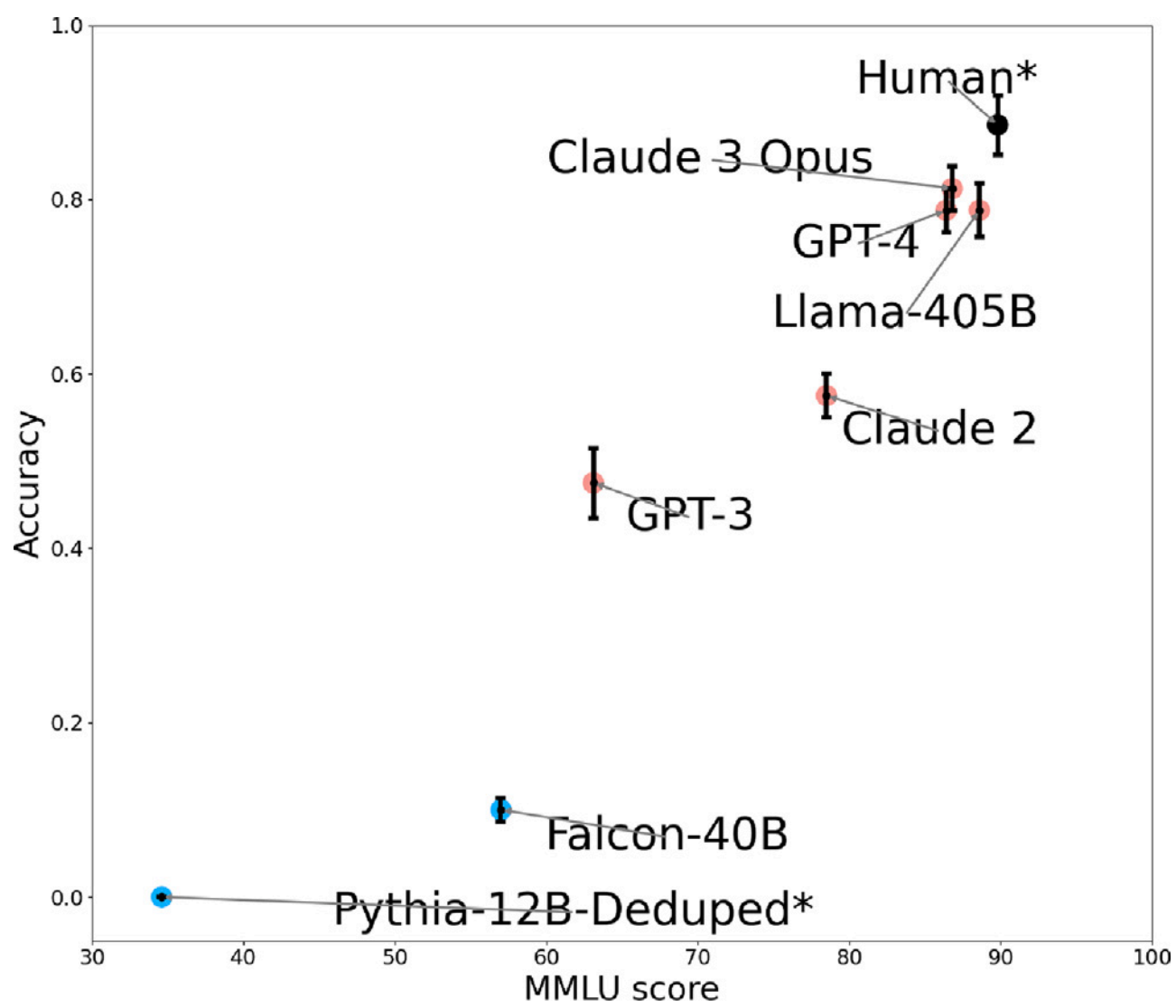

**Fig. 2.** Human and LLM reference match proportion in the Defaults condition, relative to performance on the MMLU benchmark. Models in blue are not instruction-tuned while models in peach are. Error bars show standard errors. MMLU scores are collated from various sources. (For interpretation of the references to color in this figure legend, the reader is referred to the web version of this article.)

We thus focus our remaining analysis on comparing human subjects to GPT-4, Claude 3, and Llama-405B as these are the best performing models. In the Defaults condition, none of GPT-4 (coef = −0.7696, z = −1.659, p = 0.097), Claude 3 (coef = −0.6131, z = −1.299, p = 0.194), or Llama-405B (coef = −0.7696, z = −1.659, p = 0.097) performs significantly worse than human subjects. These top three models are notably similar in both their MMLU score and their corresponding performance in the Defaults condition of our experiment.

#### *Sensitivity to question presentation*

We next compare performance in Defaults to performance in our Permuted Questions, Permuted Pairs, and Distracted settings. Recall that these conditions are designed to probe sensitivity to the order and presentation of the analogy. If humans and models exhibit different changes in performance relative to Defaults in these conditions, this would be indicative of differences in the mechanisms being used to infer the analogy from the provided text.

Fig. 3 compares humans to high-performing LLMs in the Defaults and Permuted Pairs conditions. LLM performance drops in the Permuted Pairs condition, while humans seem equally able to infer the mapping regardless of word presentation order. This effect is significant for Claude 3 (coef = −1.7802, z = −4.217, $p < 0.001$), GPT-4 (coef = −1.6796, z = −3.975, $p < 0.001$), and Llama-405B (coef = −2.3381, z = −5.500, $p < 0.001$). The remaining control conditions and data for all tested models are shown in Fig. A.15 of Appendix. In these conditions, we find that humans and models are roughly equally affected. For example, the proportion of matches with the reference answers in the Distracted condition drops by approximately 0.25 for all four subject types.

#### *Effect of Semantic Structure on reasoning*

We next investigate more directly the extent to which humans and LLMs leverage semantic structure in order to complete our analogy tasks by comparing performance in Defaults to performance in the Only RHS, Randoms, and Random Finals conditions. Recall (from Section “Study 1: Analogical inference from Semantic Structure”) that we interpret high performance in the Only RHS condition as evidence that a subject is able to complete the questions based only on the pattern in the target domain. We take the performance difference between the Only RHS condition and either the Randoms or Random Finals condition to be a measure of the subject's bias toward using the semantic structure: a large performance degradation indicates that the subject preferred to consider the LHS as meaningful even when ignoring the LHS entirely would have arguably afforded a simpler strategy for producing responses.

[3] We use “match to reference” instead of “accuracy” as there is, by definition, room for interpretation within analogy making and therefore no objectively correct solutions. For analysis purposes, we grade responses against specific reference answers. We note that, in principle, since subjects receive 4 analogy questions per quiz, the reference solution should be preferred over other justifiable solutions after seeing multiple examples.

[4] MMLU scores are few-shot for GPT-4 and 5-shot for other models. The reported human baseline is the estimate for human experts given by Hendrycks et al. (2021). The score for Pythia 12B could not be found and so we use the reported value for Pythia 6.9B Tulu.

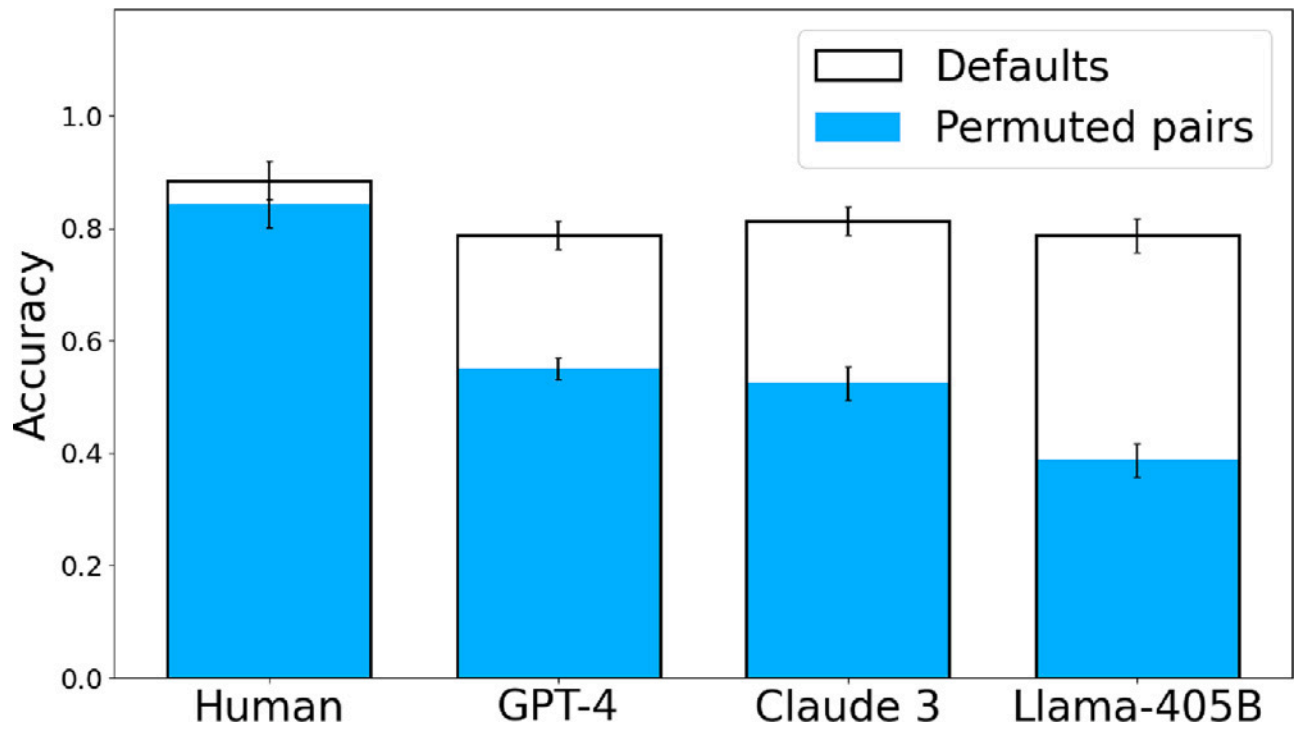

**Fig. 3.** Human and LLM reference match proportion in the Defaults and Permuted Pairs conditions. Error bars show standard errors.

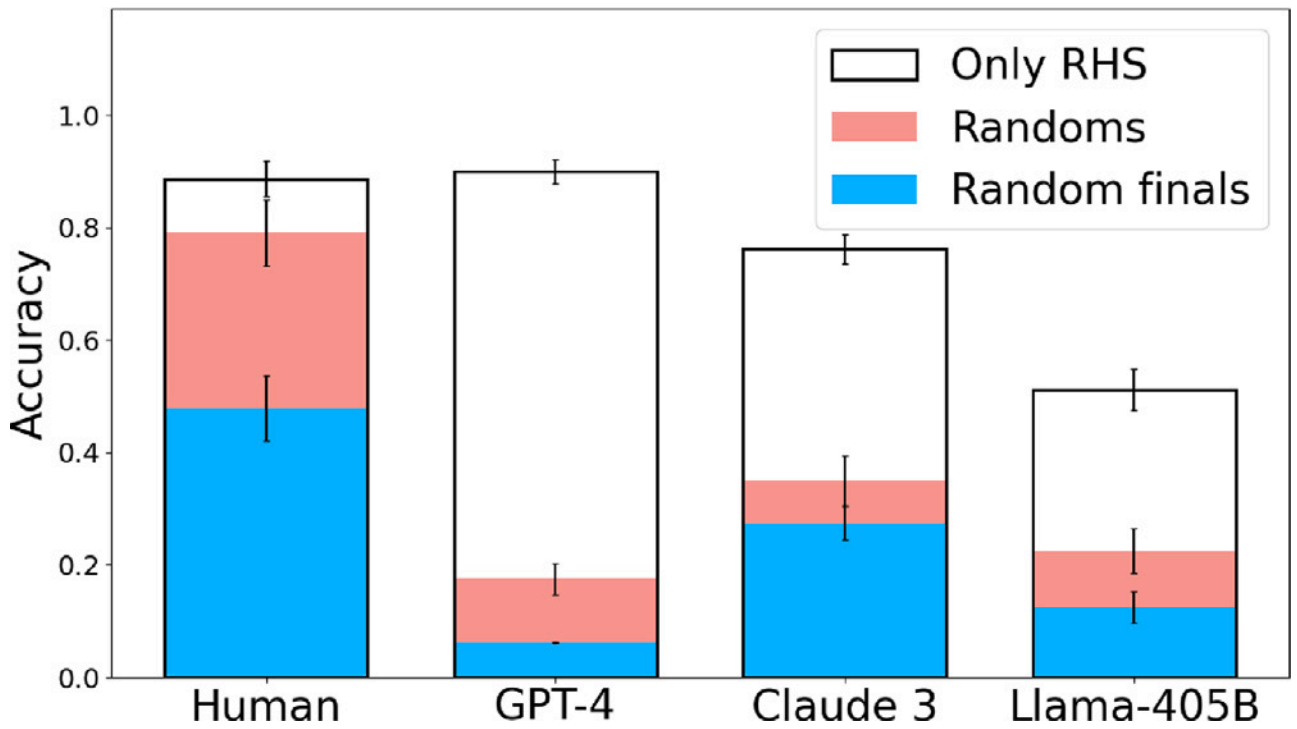

**Fig. 4.** Human and LLM reference match proportion in Only RHS, Randoms, and Random Finals conditions. Data from the Random Permuted Pairs condition is shown in Fig. A.15 of Appendix. Error bars show standard errors.

Both humans and some models are readily able to infer the reference answer in the Only RHS condition, as shown in Fig. 4. Match to reference answers is approximately 0.8 for Claude 3 with human subjects and GPT-4 slightly higher at 0.9. Llama-405B achieves lower but still reasonable matches to the reference of approximately 0.5. GPT-4 is not significantly different from humans in this condition (coef = 0.1178, z = 0.223, p = 0.824), while Claude 3 (coef = −0.9130, z = −1.994, p = 0.046) and Llama-405B (coef = −2.0294, z = −4.648, $p < 0.001$) are significantly worse than human subjects. Thus, both humans and some LLMs are able to complete the task without the guidance of the left-hand side.

Considering this, we next look at the performance degradation associated with encountering incoherent semantic structure on the left-hand side. Humans exhibit a modest decrease in matches to the reference answers of about 0.15 in the Randoms and Random Permuted Pairs conditions relative to Defaults. Claude-3 and GPT-4, however, exhibit much larger drops: Claude 3 decreases by approximately 0.5 relative to Defaults, while GPT-4 decreases by 0.6 and 0.4 in the Randoms and Random Permuted Pairs conditions. The performance of Llama-405B is comparably low in these two conditions, although reducing from a lower score in the Only RHS condition. Across these two conditions, GPT-4 (coef = −2.1972, z = −5.211, $p < 0.001$), Claude 3 (coef = −2.0680, z = −4.960, $p < 0.001$), and Llama-405B (coef = −2.7383, z = −6.078, $p < 0.001$) perform significantly worse than humans.

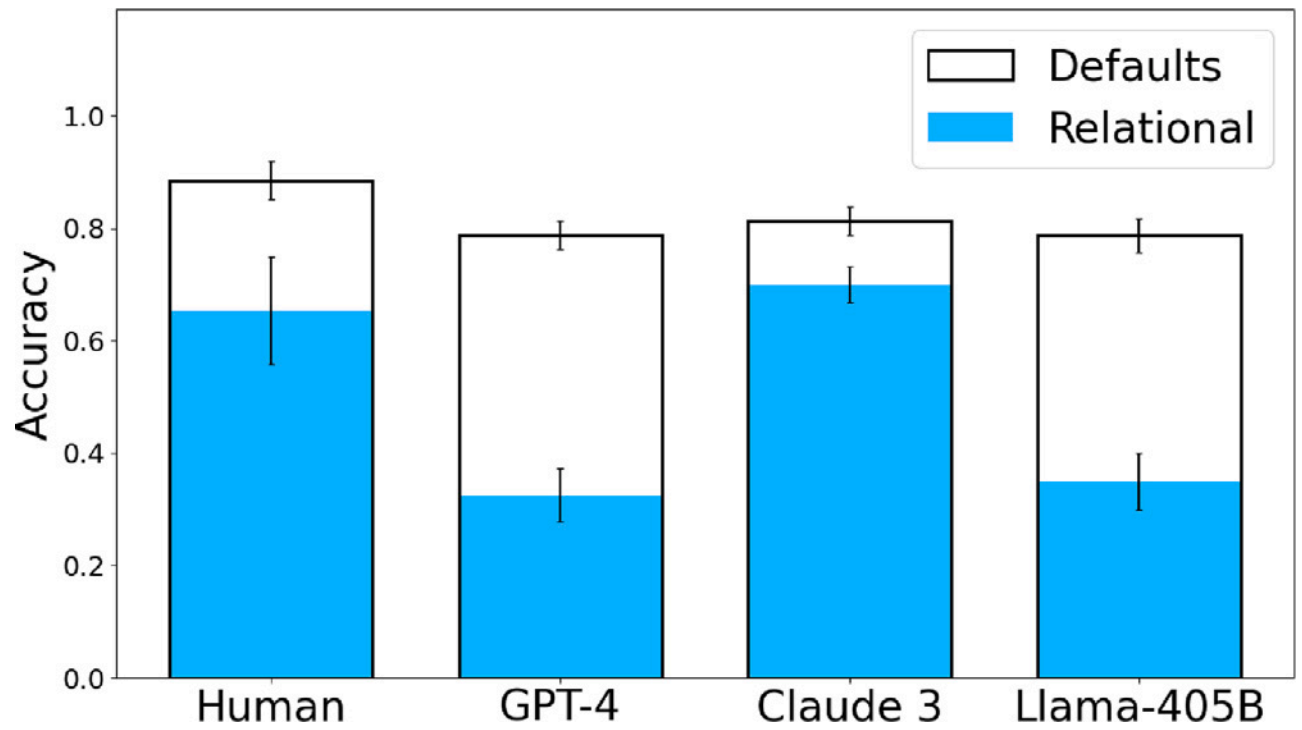

**Fig. 5.** Human and LLM reference match proportion in the $2 \times n$ condition followup, with Defaults condition performance for reference. Error bars show standard errors.

From this we conclude that human subjects are able to easily identify when the left-hand side contains no useful semantic structure to leverage. In such cases, they are able to employ a strategy that only relies on the right-hand side. By contrast, models do not seem capable of easily identifying the lack of informativeness of the left-hand side in these conditions, as they do not use the strategy of only attending to the right-hand side, even though they show their capability of using this strategy when no left-hand side is present. This suggests mechanistic differences between how human subjects and models process this task.

Although the performance of human subjects does not drop notably in the Randoms condition compared to the Only RHS condition, it does drop by a wide margin in the Random Finals condition. In this condition, correspondence with reference answers is approximately 0.5 lower than in the Only RHS condition. This further suggests that the semantic relatedness of the left-hand side affects the strategy of human subjects: when the left-hand side is clearly unrelated, the information it provides is discarded, but when much of the left-hand side appears related, the information is not discarded and the random final word of the source domain prompts an incorrect answer from human subjects. Models also show a large drop in performance in the Random Finals condition relative to Only RHS, with Claude 3 dropping by 0.5, GPT-4 dropping by 0.8, and Llama-405B dropping by 0.4 (with this smaller drop being due to a lower starting point of 0.5). Simple effects analysis shows that Claude 3 (coef = −1.0464, z = −2.799, p = 0.005), GPT-4 (coef = −2.7850, z = −5.168, p < 0.001), and Llama-405B (coef = −2.0229, z = −4.625, p < 0.001) are significantly worse than humans in the Random Finals condition. However, we see this difference as less informative given that all models drop in performance across all the random conditions relative to their own performance in the Only RHS condition.

*Diagnosing the use of an RHS-only heuristic*

To clarify whether subjects actually make use of left–right relations or only complete right-side patterns in the Semantic Structure experiment, we use our $2 \times n$ variant of the Defaults condition, preventing the use of only right-hand terms.

Results are shown in Fig. 5. Human subjects and Claude 3 exhibit similar performance, with accuracies of approximately 0.7. GPT-4 and Llama-405B, however, both attain a much lower reference match ratio close to 0.35. Simple effects analysis shows that GPT-4 (coef = −1.3669, z = −3.065, p = 0.002) and Llama-405B (coef = −1.2550, z = −2.843, p = 0.004) obtain significantly worse match to the reference answers than human subjects, while the reference match of Claude 3 does not differ significantly from human subjects (coef = 0.2111, z = 0.467, p = 0.640).

*Other observations*

We additionally analyze the extent to which human subjects and models improve with each question (Fig. A.14 of the Appendix), and the extent to which the errors made by humans and models follow the same distribution across questions grouped by target domain and across qualitative error types (Fig. A.13 and Table A.5 of the Appendix). We find that humans and models alike improve over subsequent questions, adding to a body of evidence about in-context learning (Raventós, Paul, Chen, & Ganguli, 2023; Xie, Raghunathan, Liang, & Ma, 2022; Zhang, Zhang, Yang, & Wang, 2023). Humans and models show similar error distributions by target domain, but qualitative error types reveal a closer correspondence between human and GPT-4 errors than Claude 3.

*Takeaways*

Despite weak performance from many models on our analogical reasoning tasks, GPT-4, Claude 3, and Llama-405B perform well, showing similar patterns to humans in leveraging semantic structure of corresponding domains to solve analogies. However, differences remain in how they handle semantic structure in the source domain. Humans prefer leveraging semantic structure when a clear pattern exists (evidenced by the Defaults and Random Finals conditions) but can ignore words when structure is lacking (Randoms condition). Models show the former bias but not the latter ability, appearing distracted by random lexical items. Nevertheless, newer models increasingly resemble human subjects, suggesting larger models may close this gap.

Furthermore, qualitative differences exist even between the best models. GPT-4, Claude 3, and Llama-405B match human performance in the Defaults condition, but when the structure is generalized from $2 \times 2$ to the $2 \times n$ followup, making a right-hand-only strategy unworkable, Claude 3 maintains human-level performance while GPT-4 and Llama-405B drop significantly. Despite limited public information, it is notable that models produced using presumably similar approaches can exhibit meaningfully different behavioral patterns.

*Reasoning from Semantic Content*

The above Semantic Structure experiments provide insight into the relative bias of human subjects and models to transfer this structure across domains. The Semantic Content experiments modify the tasks to investigate the extent to which human subjects and models can transfer elements of the linguistic meaning of terms from one domain to another as elements of the target domain in this experiment directly depend on properties of corresponding source domain elements, requiring knowledge of the source domain terms' meaning for perfect performance. We again present overall performance observations for humans and LLMs first (Section "Human performance continues to be robust"), before providing more specific insights about how individual LLMs differ in their performance across conditions (Section "LLM-specific variations") and about how humans vs. LLMs respond to compositional reasoning tasks (Section "Effect of compositionality").

*Human performance continues to be robust*

Results for human subjects, GPT-4, and Claude 3 are shown in Fig. 6. The other models we tested attain a much lower match rate as before, and thus are omitted from our primary analysis. Human match percentages range from 0.4 to 0.8 across conditions, comparable to the earlier Semantic Structure experiments. In qualitative analysis, subjects generally describe their strategy as relating properties of the left-hand terms to their representations on the right-hand side. For example, one subject describes their strategy as "Looking at other comparable entries to figure out what the answer to the last entry was (dog:puppy::cat:kitten)", and reporting a strategy in which "I tried to find similar pairs of entries and looked at their meanings". Appendix "Further details of human performance" contains further qualitative analysis.

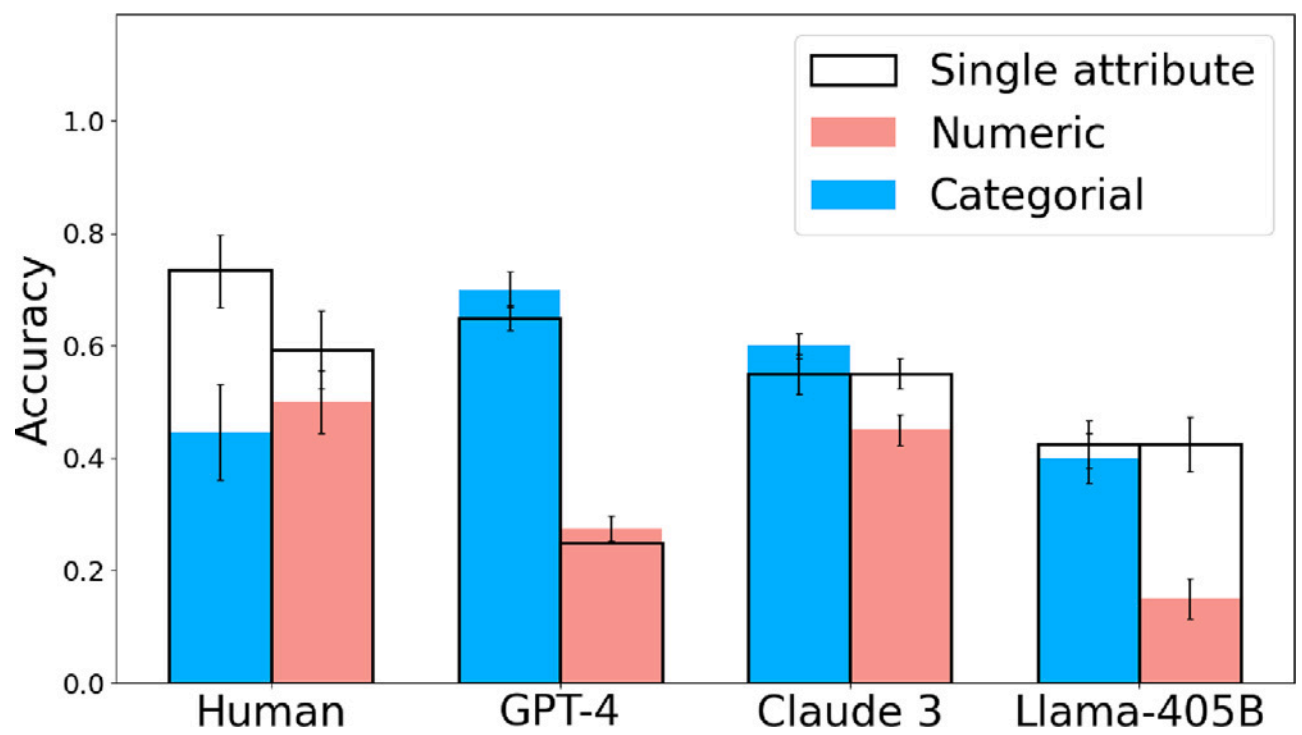

**Fig. 6.** Human and model reference match proportion by condition in the Semantic Content experiment. Error bars show standard errors. Outline bars correspond to single-attribute conditions while filled bars correspond to dual-attribute conditions, with color differentiating whether numeric properties or only categorial properties are included. For example, performance in the single-attribute Numeric condition is given by the outlines surrounding peach bars while the peach bars give performance in the Numeric Dual-attribute condition. Note that "Categorial" conditions include one or two categorial properties while "Numeric" conditions include either a numeric property alone or a numeric property paired with a categorial one. (For interpretation of the references to color in this figure legend, the reader is referred to the web version of this article.)

*LLM-specific variations*

It is worth noting that the individual LLMs do not necessarily pattern alike. We describe several interesting idiosyncrasies below.

Claude 3 matches human performance stably across the different conditions of the Semantic Content experiment with its match rate falling into a comparable range of 0.4 to 0.7. The model exhibits marginally better performance in the Dual-attribute condition and marginally worse performance in the remaining three. These differences are insignificant across all conditions, which covers the Categorial (coef = −0.8109, z = −1.879, p = 0.060), Dual-attribute (coef = 0.6206, z = 1.478, p = 0.140), Numeric (coef = −0.1788, z = −0.439, p = 0.661), and Numeric Dual-attribute (coef = −0.2009, z = −0.484, p = 0.629) conditions. In summary, Claude 3 performs as well as human subjects across all conditions of this experiment.

GPT-4 achieves good results in the Categorial and Dual-attribute conditions, with mean match proportions of approximately 0.7 in both (compared to 0.7 and 0.4 respectively for human subjects). GPT-4 is not significantly worse than humans in the Categorial condition (coef = −0.3927, z = −0.889, p = 0.374), and it significantly outperforms human subjects in the Dual-attribute condition (coef = 1.0624, z = 2.429, p = 0.015). However, its match to reference rate drops to 0.2–0.3 in the remaining conditions and we find that GPT-4 is significantly worse than humans in both the Numeric (coef = −1.4781, z = −3.321, p = 0.001) and Numeric Dual-attribute conditions (coef = −0.9694, z = −2.185, p = 0.029).

In these conditions, GPT-4 fails to correctly relate the number of characters in a response to the numeric property of the object (see Table A.7 for an illustrative example). GPT-4's failure to reason about the number of characters in the expected way is further observed in Table A.8 of the Appendix, even when the model is not required to relate a property of a word to its representation.

Llama-405B achieves very consistent results of 0.4–0.5 in three conditions but a lower score of approximately 0.2 in the Numeric Dual-attribute condition. Performance is significantly lower than human subjects in the Categorial condition (coef = −1.3139, z = −3.034, p = 0.002), does not differ significantly from human subjects in the Dual-attribute (coef = −0.1904, z = −0.453, p = 0.650) or Numeric conditions (coef = −0.6818, z = −1.668, p = 0.095), and is significantly worse than human subjects in the Numeric Dual-attribute condition (coef = −1.7346, z = −3.354, p = 0.001). Llama-405B thus appears to struggle to combine numeric reasoning with reasoning about two properties simultaneously in our experiments, despite its ability to reason about two non-numeric properties or about a single numeric property.

*Effect of compositionality*

When comparing the performance of a subject in a non-compositional (single-attribute) condition to the corresponding compositional (dual-attribute) version, we observe some decrease in performance for human subjects but not for models. This admittedly unexpected result has various possible explanations, which we address in the discussion below. The performance of human subjects drops from approximately 0.7 to approximately 0.4 when comparing the Categorial condition to the corresponding compositional version (the Dual-attribute condition). A simple effects analysis confirms that this decline is significant (coef = −1.2267, z = −3.091, p = 0.002). We see a non-significant decrease in performance for human subjects when comparing the Numeric condition to its compositional counterpart, with performance dropping from approximately 0.6 to approximately 0.5 (coef = −0.3795, z = −1.028, p = 0.304).

By contrast, we do not find the models to be significantly worse in compositional conditions than non-compositional ones. In fact, GPT-4 exhibits a slight improvement in the compositional conditions, though this change is statistically insignificant for both the Dual-attribute condition relative to the Categorial condition (coef = 0.2281, z = 0.477, p = 0.634) and for the Numeric Dual-attribute condition relative to the Numeric condition (coef = 0.1292, z = 0.254, p = 0.799). For Claude 3, we similarly find the differences to be insignificant for the Dual-attribute condition relative to the Categorial condition (coef = 0.2049, z = 0.452, p = 0.651) and for the Numeric Dual-attribute condition relative to the Numeric condition (coef = −0.4013, z = −0.893, p = 0.372). Llama-405B is insignificantly worse in the Dual-attribute condition relative to the Categorial condition (coef = −0.1030, z = −0.227, p = 0.821), and its anomalously low performance in the Numeric Dual-attribute condition yields a significant difference in that condition relative to the Numeric condition (coef = −1.4323, z = −2.622, p = 0.009).

*Takeaways*

The Semantic Content experiment confirms that human subjects perform robustly and flexibly across diverse task variations. Claude 3 matches human performance in all conditions, indicating it shares humans' tendency to use the source domain's semantic content when completing target domains. While GPT-4's poor performance in numeric conditions is notable, it most likely reflects a failure in numeric reasoning rather than a difference in analogical reasoning.

We find evidence of decreased human performance, but not model performance, in compositional conditions, contrasting with some existing research showing that models may struggle with composition (Press et al., 2023). However, other factors may be at play. In particular, models may indeed struggle with composition, but be aided by an increased availability of information in this condition. What appears as undiminished model performance in compositional conditions may be the result of a balanced negative and positive effect on models: a negative effect of the difficulty of composition, and a positive effect of increased information availability (in the compositional conditions the target domain represents two properties, which may allow models to more easily recognize the encoding of source domain properties). On the other hand, human subjects may show reduced performance in the compositional conditions because of the difficulty of composition, unmasked by a competing positive effect (the human subjects, unlike the models, may not struggle to determine the source-target mapping even when only one property is involved). Thus, the discrepancy between human and model performance in this condition might not be indicative of a difference in compositional capability between humans and models.

*Evaluating the effect of chain-of-thought*

We provide the best-performing model, Claude 3, with a chain-of-thought prompt across all experiments to investigate whether this improves performance. Across several prompt variations we observe weak results, documented in Appendix "Chain-of-thought prompting". Therefore, the prompting strategy reported in the main text does not appear to be uncharitable to models.

## Discussion

Our results show that the best-performing LLMs are able to successfully complete many analogical reasoning tasks in a way that is comparable to humans, even when using novel stimuli that are highly unlikely to appear in their training data. That said, behavior variation across task presentations and control conditions reveals that there remain meaningful differences in how such analogies are processed, evidenced by differences in how humans and models respond to distracting or misleading information. Nonetheless, we observe a clear trend in which more recent leading models come increasingly close to matching human performance across our tasks. In particular, Claude 3 exhibits impressively robust performance across most task variations, even closing the gap with humans in some test conditions in which the earlier GPT-4 exhibited limitations (such as the $2 \times n$ task in which mapping from the source domain must be used for success). Within our sample of models, GPT-4 and Claude 3 may have distinctive features correlated with their recency, such as size or innovations in architecture and post-training, though such features are difficult to judge given that the models are closed-source. Within open source models, the good performance of the very large Llama 405B despite a relatively straightforward design (since it is not a mixture of experts and follows a standard post-training pipeline) suggests that parameter and data scale may underlie a significant but unknown proportion of recent performance increases, although other factors are certainly involved. Together, these results raise questions about the ability of LLMs and similar models to serve as candidate cognitive models, which we discuss briefly below.

*Evaluating the competence of LLMs*

*Robust performance of state-of-the-art models*

The breadth of Claude 3's success on our tasks — which likely do not appear in its training data — is noteworthy. It suggests that state-of-the-art LLMs can broadly match human performance not only in formal analogical reasoning tasks but also in tasks that require re-representing and transferring semantic information across linguistic and non-linguistic domains. As such, our results weigh against a view held by some cognitive scientists according to which connectionist models without a built-in symbolic component are constitutively limited in their ability to robustly handle analogical reasoning tasks. This view is bolstered by a long history of only modest progress in analogical reasoning using deep learning systems, coupled with criticisms of their limited successes when tested on truly held out test sets (Mitchell, 2021). They also inform discussions of whether LLMs possess "functional" linguistic competence, in addition to "formal" linguistic competence (Mahowald et al., 2023). Further work is needed to characterize the precise mechanism that LLMs are using to solve these tasks; it is possible — though increasingly unlikely given the robustness of the behavioral results — that success is due to a myriad of heuristics rather than a systematic analogical reasoning process. Even so, evidence of LLMs completing analogical reasoning tasks in domains designed to involve relating semantic information to non-linguistic correlates, in addition to tasks over abstract symbols, supports the claim that some LLMs are capable of functional linguistic competence in addition to formal competence. Such an outcome would be consistent with the argument made in Mahowald et al. (2023), which suggested that such capabilities might emerge if models were exposed to particular types of post-training, as all successful models in are study were.

*Where today's models still stumble*

There remain examples of LLMs performing much worse than humans on analogical reasoning tasks (Lewis & Mitchell, 2024), which must be reconciled with our results. Here the competence-performance distinction, originally introduced by Chomsky (1965), can be usefully applied to the evaluation of LLMs (Firestone, 2020; Pavlick, 2022, 2023). This distinction allows researchers to theorize about the abstract computational principles governing cognition independently from other factors that introduce "noise" into observed behavior. In humans, it is generally assumed that there is a double dissociation between performance and competence: neither success nor failure on a task designed to measure a particular capacity can always be taken as conclusive evidence that subjects have or lack that capacity, due to auxiliary factors affecting task performance. When it comes to LLMs, by contrast, the distinction is typically applied in a single direction: human-like performance on benchmarks is often explained away by reliance on shallow heuristics (McCoy, Yao, Friedman, Hardy, & Griffiths, 2024) and/or lack of construct validity (Ullman, 2023), while sub-human performance is often taken as reliable evidence of lack of competence. However, LLM performance can also be negatively affected by strong auxiliary task demands (Hu & Frank, 2024) and mismatched conditions in comparisons with human subjects (Lampinen, 2023). These are compelling reasons to apply the dissociation in both directions to LLMs as well.

*Competence vs. Performance: A two-way street*

From this perspective, our results offer evidence to support both sides of the present debate about whether LLMs possess human-level analogical reasoning (see Hodel and West (2024), Lewis and Mitchell (2024), and Webb et al. (2024)). Supporting the argument of Webb et al. (2024) that deficiencies in capabilities other than analogical reasoning can explain poor model performance in some tasks, we find that GPT-4's failure in the numeric conditions of our Semantic Content experiment may be due to a deficiency in counting ability. However, contrary to Webb et al. (2023), who report impressive analogical reasoning in both GPT-3 and GPT-4, we do find a notable difference in the performance of these two models, with GPT-3 performing quite poorly on our tasks. Among the models tested, only GPT-4, Claude 3, and Llama-405B produce results that merit detailed comparison with human subjects. This suggests that claims of human-level performance of LLMs on analogical reasoning tasks may have been premature and might have relied on insufficiently challenging tasks (although note that Webb et al. (2023) test GPT-3 text-davinci-003 while we test GPT-3 text-davinci-002).

However, other differences we observe between human subjects and LLMs across task variations are not subject to an auxiliary task demand explanation and suggest that the underlying mechanisms of analogical reasoning in these systems may differ from that in humans. Importantly, these differences persist even in our best performing model, Claude 3. For instance, Claude 3 responds differently than human subjects when some or all words in the source domain are replaced with random words, indicating that they may use distinct strategies for identifying and leveraging relational similarities between source and target domains. Furthermore, Claude 3 remains more sensitive than human subjects to the ordering of elements within domains, which is difficult to explain if LLMs are using a generalizable symbolic working memory approach.

Collectively, these patterns bear on the larger question of how we should arbitrate disputes about competence in machine-human comparisons. On the one hand, it seems reasonable to assume that any system that can reliably achieve success at or above human levels on experiments like ours — without relying on memorization and other confounds — should be considered competent at analogical reasoning. On the other hand, we should be open to the possibility that such competence may be implemented differently in LLMs and humans (Millière & Rathkopf, 2025).

The question of whether we require human-likeness of a mechanism to declare human-level "competence" is ultimately not empirical, but rather demands philosophical consensus among the scientific community around our ultimate goals and metrics for achieving them.

*Can LLMs support progress in understanding analogical reasoning?*

*What analogical reasoning in LLMs can tell us*

A central question for our study and similar research is: what is the relevance of LLMs' analogical reasoning abilities to human cognitive theory? By designing a set of analogical reasoning tasks which are not readily explained by existing theories and models of analogy, we demonstrate that LLMs, at a minimum, expand the empirical coverage of computational models of analogy. This finding alone necessitates that they be taken seriously, and that their mechanisms be investigated further, before their significance to human analogical reasoning is dismissed.

More specifically, we can consider how our tasks relate to one of the leading theories of analogical reasoning, Structure Mapping Theory (SMT). A task that is directly covered by SMT involves establishing the correspondence between two domains based on aligning higher-order relations that ignore the semantic content of the entities of each domain. By contrast, our experiment (1) involves semantic content in one of the two experiments, (2) requires subjects to make an analogical inference by performing rule induction (Rule et al., 2018) or schema induction (Gick & Holyoak, 1983), and (3) requires the flexible re-representation of object terms so as to focus on the analogy-relevant properties of the domains. Each of these three aspects corresponds to an open set of theoretical problems in understanding analogical reasoning.

Incorporating semantic information into our analogical reasoning problems ensures that we investigate whether LLMs' analogical reasoning resembles that of humans in a relevant way to its purportedly central role in broader cognition. Following Gentner (1983), emphasis has been placed on relational similarity, rather than just feature similarity, in mapping from a familiar source to a foreign target domain during analogical reasoning to allow for the flexible transfer of knowledge (Gust, Krumnack, Kühnberger, & Schwering, 2008; Halford, Wilson, & Phillips, 2010; Holyoak, 2012). This conception allows analogical reasoning to play a fundamental role in human cognition, supporting the emergence of diverse cognitive abilities via "bootstrapping" (Carey, 2004, 2009; Gentner, 2010). In bootstrapping, two cognitive processes mutually support each other's development. In Gentner's Structure-Mapping Theory (SMT), linguistic competence and structure-mapping-based analogical reasoning are hypothesized to co-develop, with structure-mapping supporting the necessary relational reasoning to model language-world relations, and language acquisition in turn developing symbolic reasoning capacities that amplify structure-mapping abilities. Consequently, analogical reasoning is seen as a central cognitive phenomenon of interest.

Implicitly requiring subjects in our experiments to infer an abstract rule or schema that governs the relationship between a source and target domain, and to do so based on flexibly re-representing inputs in a suitable way to support this rule discovery, addresses the core of an open problem in analogical reasoning. As Chalmers et al. argue, analogical reasoning over inputs that are already represented in a way that supports such reasoning can be trivial and not informative of analogical inputs over unconstrained natural inputs: problem-specific representations hand-crafted by researchers can offload a significant portion of the reasoning problem (Chalmers et al., 1992). Before a subject can search over a large space of possible rules governing the correspondence between two domains, they must first search over a large space of possible ways of representing the entities in the domains, only some of which correspond to simple rules relating the domains. This feature-selection process requires non-trivial reasoning about which properties of the system are relevant in the given context.

The requirement of our tasks to perform rule or schema induction based on the re-representation of real-world inputs is important for understanding analogical reasoning's role in natural contexts. Current theories either address re-representation without real-world concepts, or handle real-world concepts without tackling re-representation. For example, Mitchell and Hofstadter's Copycat models (Hofstadter & Mitchell, 1995) addressed re-representation but in the domain of letter string analogies, while subsequent work addressed re-representation in a synthetic Raven's matrices domain (Lovett & Forbus, 2017). Conversely, recent efforts by Lu, Ichien, and Holyoak (2021) and Lu et al. (2019) address real-world inputs but neglect the problem of re-representation.

*Toward mechanistic insight and future work*

The success of some LLMs in many of our tasks suggests that the most advanced models may be capable of employing an analogical reasoning approach that captures some aspects of human analogical reasoning in a domain that is not well-covered by existing theory. The potential emergence of such competence from training primarily on text prediction could yield new hypotheses about the emergence of analogical reasoning as a central cognitive faculty from generic learning mechanisms (possibly combined with the unique pressures of language acquisition).[5] However, matching human behavioral performance with black-box models only provides limited explanatory insight into human analogical reasoning. To move beyond behavioral simulation, we should seek to establish explicit *model-to-mechanism mappings*—identifying which internal representations or computational processes in the language model correspond to the theoretical constructs posited in human analogical reasoning, examining whether the model's learning trajectory parallels human developmental patterns, and determining through mechanistic interpretability whether the model employs strategies that align with or diverge from those used by humans (Cao & Yamins, 2024; Frank & Goodman, 2025). Without such mappings, our results demonstrate impressive predictive capability but stop short of explaining how humans actually perform analogical reasoning.

The proprietary nature of leading LLMs like Claude 3 prevents this kind of investigation, but open-weights models like Llama-405B come tantalizingly close to replicating the performance of Claude 3 on our tasks. Using causal intervention methods, we could in principle reverse-engineer the algorithm implemented by a high-performing open-weights language model to solve our tasks (Geiger et al., 2025). Preliminary findings have started shedding light on the emergent symbolic mechanisms through which language models solve letter string and verbal analogies (Yang et al., 2025). Applying similar methods with our tasks could produce *how-possibly* explanations for the flexible re-representation that underlies analogical reasoning in humans.[6] In a field where available theories are partial, the ability to generate how-possibly mechanistic explanations from empirically successful models would mark a substantial advance, allowing future interpretability and experimental work to refine our cognitive theories of analogy.

That said, the mixed success of LLMs and the significant differences from humans in certain conditions underscore the need for continued research to test the robustness of any conclusion that analogical reasoning in LLMs closely matches that of human subjects. As LLM outputs continue to converge toward human responses — an expected result of the language modeling objective — it is crucial to develop novel tasks that examine analogical reasoning abilities and are almost certainly not attested in the training data. While our tasks allow for clear discrimination between human performance and that of most models prior to Claude 3, further differences in analogical reasoning patterns between humans and Claude 3 likely exist beyond those revealed by our tests. More granular testing would help clarify the extent of the remaining discrepancies between humans and the most advanced LLMs.

[5] While we do not know the proportion of simple text prediction in the training of closed models such as GPT-4 and Claude 3, the bulk of the relatively-successful Llama 405B's training appears to be text prediction. The model is pre-trained on 15.6 trillion text tokens, while the scale of post-training appears significantly smaller (see sections 4.2 and 4.3 of Grattafiori et al. (2024)). That said, Llama 405B's training being mostly text prediction does not guarantee that this is what underlies the development of its analogical reasoning capabilities.

[6] How-possibly explanations explain what could possibly cause a phenomenon, in contrast with how-actually explanations, which explain what actually causes it; with additional evidence, the former can graduate to the latter (Craver, 2006).

**CRediT authorship contribution statement**

**Sam Musker:** Writing – review & editing, Writing – original draft, Visualization, Validation, Software, Resources, Project administration, Methodology, Investigation, Formal analysis, Data curation, Conceptualization. **Alex Duchnowski:** Writing – review & editing, Writing – original draft, Visualization, Validation, Software, Resources, Project administration, Methodology, Investigation, Formal analysis, Data curation, Conceptualization. **Raphaël Millière:** Writing – review & editing, Writing – original draft, Visualization, Validation, Supervision, Software, Resources, Project administration, Methodology, Investigation, Formal analysis, Data curation, Conceptualization. **Ellie Pavlick:** Writing – review & editing, Writing – original draft, Visualization, Validation, Supervision, Resources, Project administration, Methodology, Investigation, Funding acquisition, Formal analysis, Data curation, Conceptualization.

**Declaration of competing interest**

We have no competing interests to declare.

**Acknowledgments**

We thank Taylor Webb and Melanie Mitchell for formative feedback on an earlier draft, which substantially influenced the framing and presentation of the contribution. Thank you also to Roman Feiman, Kyle Mahowald, and to the anonymous reviewers for their thoughtful and constructive input. The project depicted is sponsored in part by a Young Faculty Award from the Defense Advanced Research Projects Agency, United States, Grant #D24AP00261. The content of the information does not necessarily reflect the position, or the policy of the government and no official endorsement of this work should be inferred.

## Appendix. Statistical outputs and supplementary figures

### *A.1. Regression results, Semantic Structure experiment*

We perform a logistic regression with the outcome variable being the raw score (a 0 or 1 for each question). The predictor variables are condition and subject type (restricted to human subjects and GPT-4 only, or human subjects and Claude 3 only). The regression is performed with and without interactions:

Without interactions:

```
smf.logit(formula=respondent_scores ~
    C(subject_type, Treatment(reference=human))
    + C(quiz_class, Treatment(reference=permuted_
    questions)),
    data=all_subjects_df,
).fit(maxiter=1000, method=bfgs)
```

With interactions:

```
smf.logit(formula=respondent_scores ~
    C(subject_type, Treatment(reference=human))
    * C(quiz_class, Treatment(reference=permuted_
    questions)),
    data=all_subjects_df,
).fit(maxiter=1000, method=bfgs)
```

The significance of including the interaction between predictors is assessed with a likelihood ratio test with the associated $p$-value calculated as follows:

`p=chi2.sf(lik_ratio, degfree)`, with 7 degrees of freedom.

where likelihood ratio in the above formula is calculated as follows:

`lik_ratio = degfree * (res_subjXclass.llf - res_subjplusclass.llf).`

In the equation above, `res_subjXclass` and `res_subjplusclass` are the regression outputs with and without interactions respectively.

For both comparisons (human subjects compared to GPT-4 and human subjects compared to Claude 3), we find a significant improvement in model fit when interactions between the subject type and experiment condition are included. A likelihood ratio test shows that including interactions between subject type and experiment condition leads to a significantly better fit of the model ($chi^2(7) = 115.1871, p < 0.001$). For the comparison between Claude 3 and human subjects, we again find a significant negative effect of the subject type being Claude 3 when interactions are not included (coef = −0.8706, z = −5.608, $p < 0.001$) and find that subject type — condition interactions are significant ($chi^2(7) = 173.6511, p < 0.001$). These results are consistent with the observation that the two models exhibit variable performance across conditions, and indicate that the overall performance gap to human subjects is driven by low model match to reference in certain conditions. Simple effects analysis is used below to assess the effect of subject type in particular conditions and groups thereof.

Regression outputs are provided in the linked Github repository.

*A.2. Regression results, semantic content experiment*

Regressions are performed in the same manner as for the Semantic Structure experiment, described in Appendix “Regression results, Semantic Structure experiment”. Here, the reference condition is Categorial and the degrees of freedom used for the likelihood ratio test is 4.

As observed in the Semantic Structure experiment, the performance of GPT-4 in the Semantic Content experiment is human-comparable in some conditions but notably lower in others. When comparing a logistic model that uses subject type and experiment condition separately to one that includes their interactions, a likelihood ratio test shows that the model with interactions fits the data significantly better ($chi^2(4) = 39.6565, p < 0.001$). For the comparison between Claude 3 and human subjects, when comparing a logistic model that uses subject type and experiment condition separately to one that includes their interactions, a likelihood ratio test shows that the model with interactions fits the data significantly better ($chi^2(4) = 11.6002, p = 0.021$).

Regression outputs are provided in the linked Github repository.

*A.3. Chain-of-thought prompting*

We test the best-performing model from our experiments, Claude 3, with a chain-of-thought prompt across all experiments. We use the following prompt:

Chain-of-thought prompt: “We are conducting an experiment on general reasoning abilities. Below we will show you various words and drawings of each, after which you will need to complete the last drawing. Respond by thinking through the problem, reasoning step by step, before ending with “The answer is:” [and then the answer]. Get to the answer well within 100 words. If there is more than one question, only give the answer to the last question. When giving the answer, respond as concisely as possible with only the last drawing”.

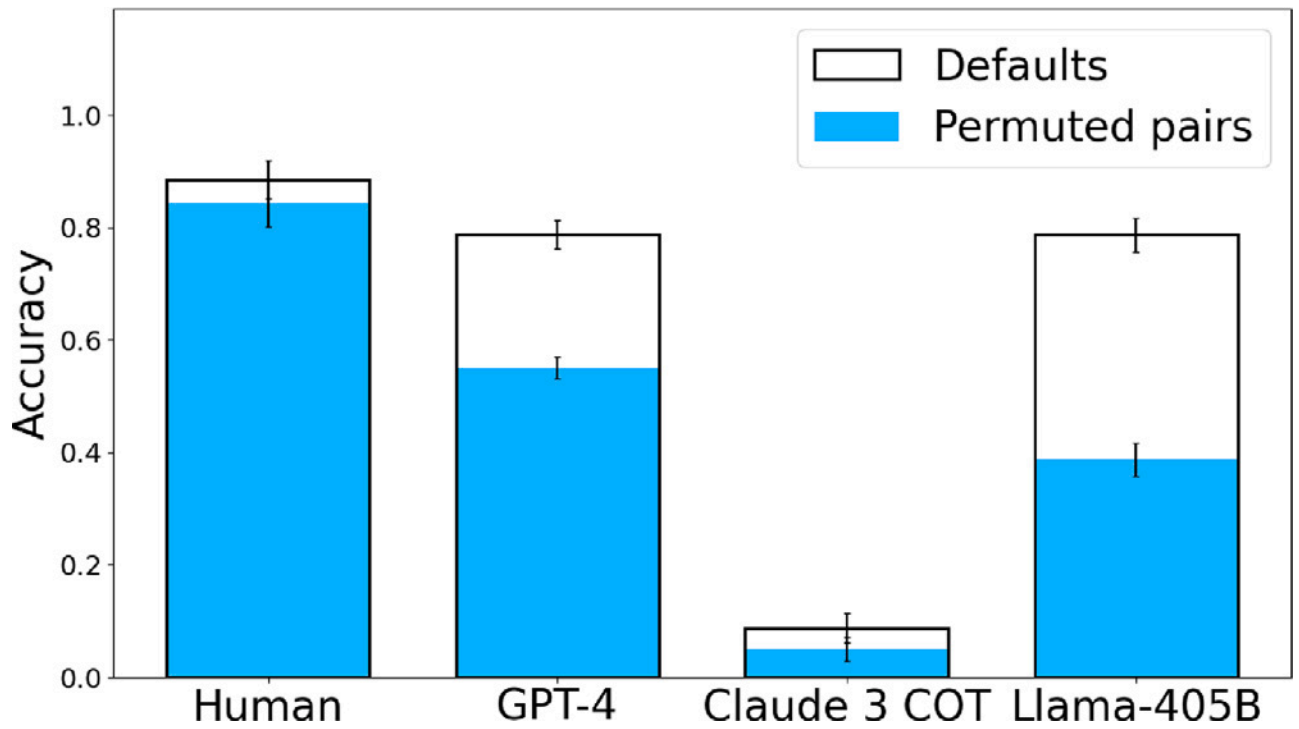

**Fig. A.7.** Human and LLM reference match proportion in the Defaults and Permuted Pairs conditions. Error bars show standard errors.

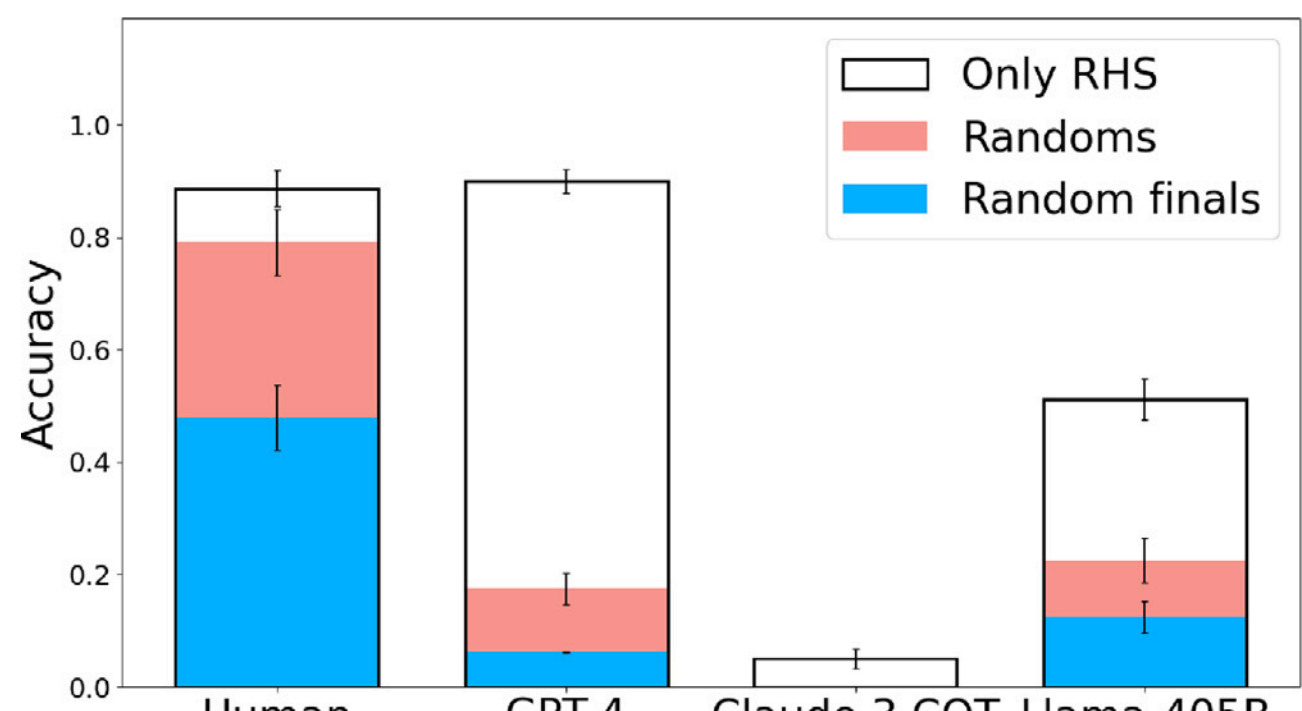

**Fig. A.8.** Human and LLM reference match proportion in Only RHS, Randoms, and Random Finals conditions. Error bars show standard errors.

We provide the model with a response length of 300 tokens, and grade the answer following “The answer is:” (when available, as is the case 97% of the time in the Semantic Structure experiment and 87% of the time in the Semantic Content experiment; when this phrase is not present we grade whatever is provided). The prompt components asking the model to (a) get to the answer within 100 words and (b) only answer the last question were added after two initial runs without these conditions resulted in the model not getting to the answer within the response window or first answering earlier questions before the live one.

The prompt yields responses that are complete and on-topic but largely wrong, with low match to reference answers across the board. This is shown in Figs. A.7, A.8, A.9, and A.10 below.

These results are surprisingly poor, and so we investigate the nature of the errors. Responses generally provide step-by-step reasoning that is related to the questions and then give an answer to the correct question, but the reasoning and answers are incorrect. An illustrative example is the following question from the Defaults condition:

Prompt: “We are conducting an experiment on general reasoning abilities. Below we will show you various words and drawings of each, after which you will need to complete the last drawing. Respond by thinking through the problem, reasoning step by step, before ending with “The answer is:” [and then the answer]. Get to the answer well within 100 words. If there is more than one question, only give the answer to the last question. When giving the answer, respond as concisely as possible with only the last drawing.

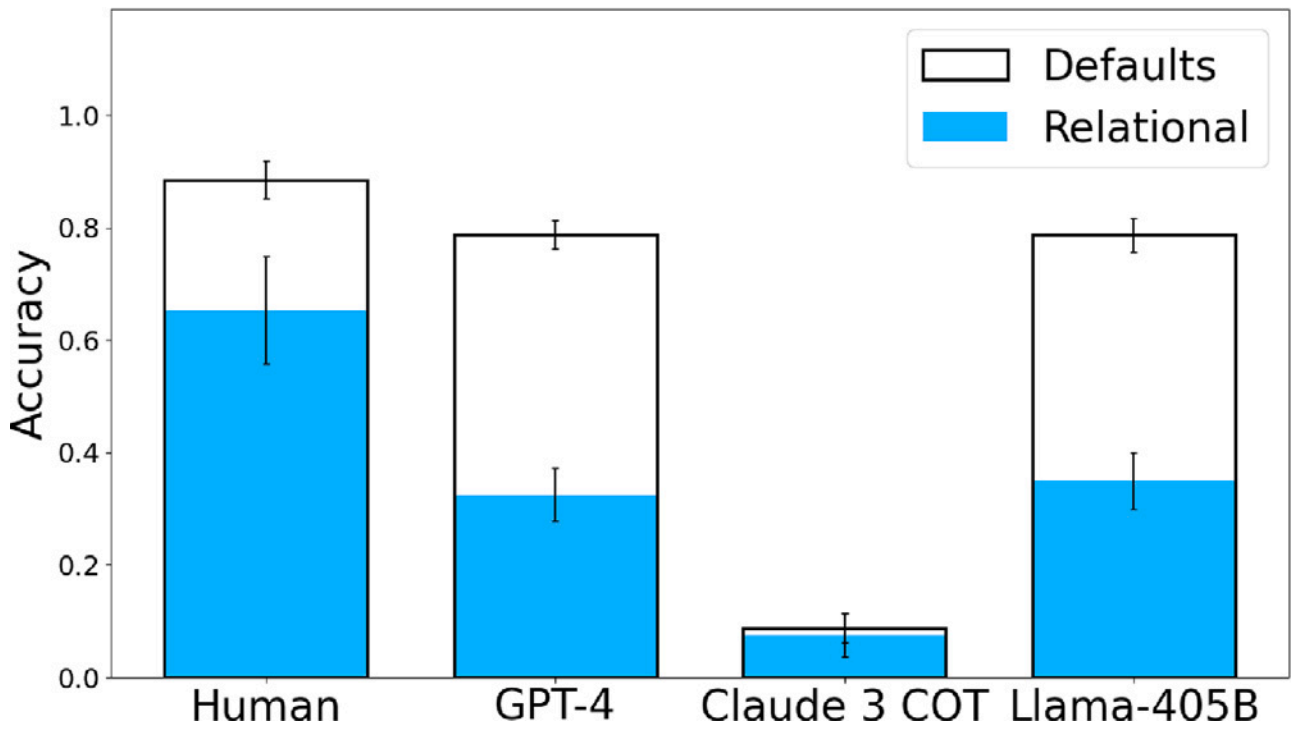

**Fig. A.9.** Human and LLM reference match proportion in the $2 \times n$ condition followup, with DEFAULTS condition performance for reference. Error bars show standard errors.

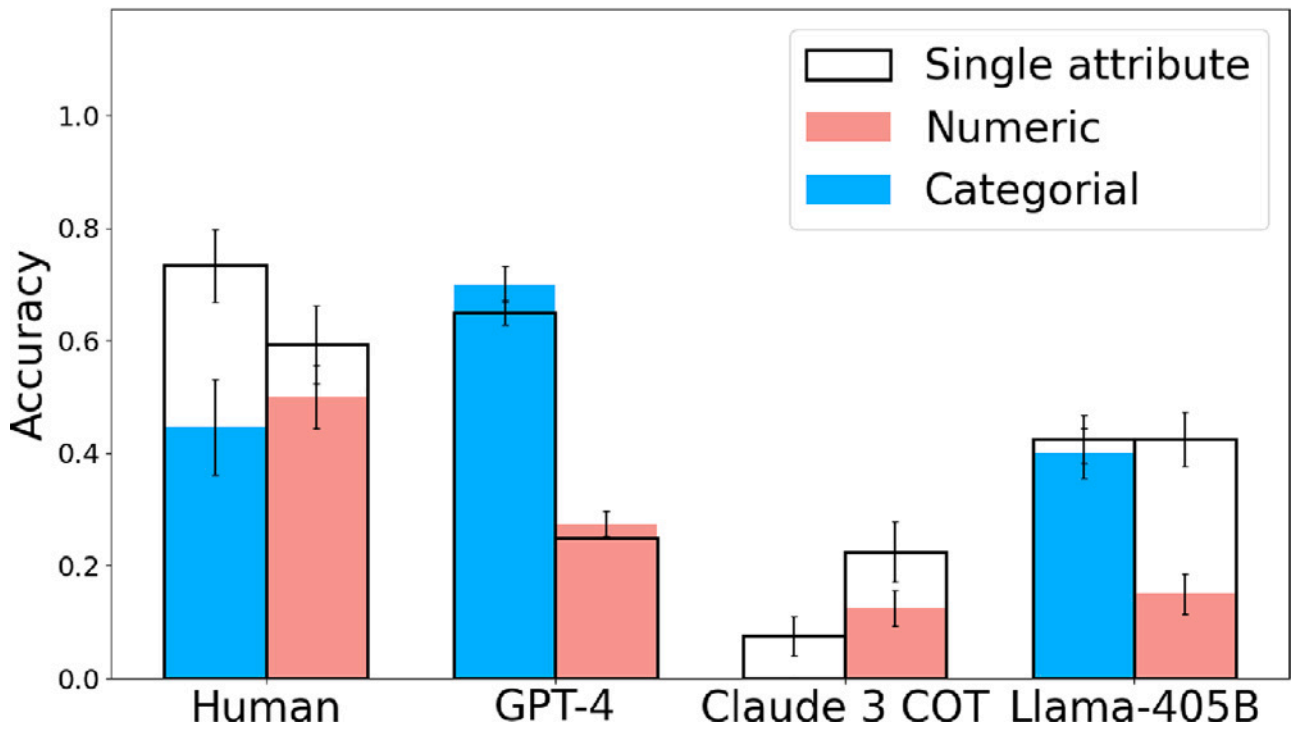

**Fig. A.10.** Human and model reference match proportion by condition in the Semantic Content experiment. Error bars show standard errors. Outline bars correspond to single-attribute conditions while filled bars correspond to dual-attribute conditions, with color differentiating whether numeric properties or only categorial properties are included. For example, performance in the single-attribute NUMERIC condition is given by the outlines surrounding peach bars while the peach bars give performance in the NUMERIC DUAL-ATTRIBUTE condition. Note that "Categorial" conditions include one or two categorial properties while "Numeric" conditions include either a numeric property alone or a numeric property paired with a categolial one. (For interpretation of the references to color in this figure legend, the reader is referred to the web version of this article.)

Question 1:
square => E K C
rectangle => E % K % C
circle => C K E
oval => "

The reference answer for this question is "C % K % E", because a rectangle is to a square as an oval is to a circle, and the transformation from square to rectangle is the insertion of percent signs.

**Model responses:**

**Response A**

"The pattern seems to be:
1. The first letter is the first letter of the shape name.
2. The middle letter is always 'K'.
3. The last letter is the last letter of the shape name.

Applying this pattern to 'oval':
1. First letter of 'oval' is 'O'.
2. Middle letter is 'K'.
3. Last letter of 'oval' is 'L'.

The answer is: O K L"

**Response B**

"The pattern seems to be:
1. The first letter represents the shape (E for rectangle, C for circle).
2. The middle letter is always K.
3. The last letter is the first letter of the shape name (C for circle, E for rectangle).

Following this pattern for an oval, we would have:
1. O for oval
2. K in the middle
3. O for the last letter

The answer is: O K O"

**Response C**

"The pattern seems to be:
1. The first letter represents the shape (square, rectangle, circle).
2. The middle letter is always K.
3. The last letter alternates between C and E, depending on whether the shape has straight sides (C) or curved sides (E).

An oval has curved sides, so it should end with E.

The answer is: O K E"

**Response D**

"The pattern seems to be:
1. The first letter represents the shape (E for rectangle, C for circle).
2. The middle letter is always K.
3. The last letter is the first letter of the shape name (C for circle, E for rectangle).

Since an oval is similar to a circle, the pattern for oval should be:

The answer is: O K O"

**Response E**

"The pattern seems to be:
1. The first letter represents the shape (E for rectangle, C for circle).
2. The middle letter is always K.
3. The last letter is the first letter of the shape name (C for circle, E for rectangle).

Applying this pattern to an oval:
1. Oval starts with O.
2. The middle letter is K.
3. The last letter is O (first letter of "oval").

The answer is: O K O"

It is interesting that Claude 3 performs well "intuitively" but worse when reasoning through the problems explicitly.

We test the recent Deepseek R1 "reasoning model" (DeepSeek-AI et al., 2025) on the same example above. Deepseek R1 is post-trained with reinforcement learning over reasoning text to improve multi-step problem solving. Interestingly and in contrast to Claude 3 with the chain-of-thought prompt, it reasons through the above prompt perfectly as shown below:

"Okay, let us try to figure out this problem. So, there are these shapes and corresponding symbols. The first one is a square which maps to E K C. Then a rectangle is E % K % C. A circle is C K E, and the question is what the oval would be.

Hmm, I need to see the pattern here. Let us list them out:
- Square: E K C
- Rectangle: E % K % C
- Circle: C K E
- Oval: ?

First, let us look at the square and rectangle since they are similar shapes. The square is a four-sided figure with equal sides, and a rectangle is a four-sided figure with opposite sides equal. The symbols for square are E K C. For rectangle, it is E % K % C. So the rectangle's symbols have % inserted between each of the square's symbols. So E becomes E %, then K becomes % K %, and C becomes % C? Wait, not exactly. Let me check again.

Wait, the square is E K C. The rectangle is E % K % C. So between each letter in the square's sequence, they added a %. So E then %, then K, then %, then C. So the square's E K C becomes E % K % C for the rectangle. That seems like inserting a % between each original character. So square to rectangle adds % between each element.

Now, circle is C K E. What would the oval be? Ovals are like stretched circles. So maybe similar to how rectangle is a stretched square. So if circle is C K E, then the oval would follow the same pattern as square to rectangle. So inserting something between the letters. But for the rectangle, they used %. Maybe for the oval, it is a different symbol? Wait, the rectangle's pattern was inserting % between each letter of the square. So square (E K C) becomes E % K % C. So each space between letters gets a %.

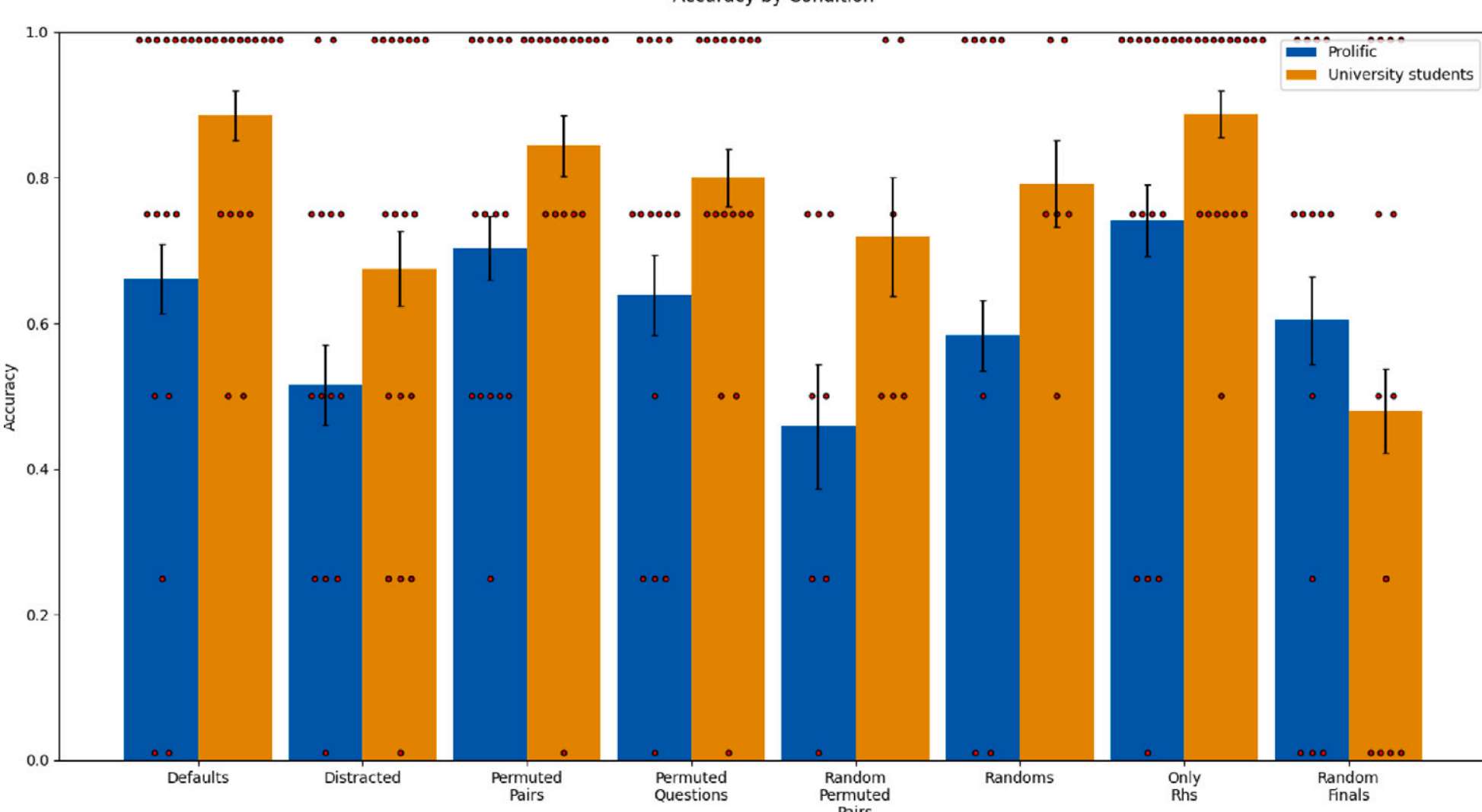

**Fig. A.11.** Proportion of match to reference answers comparison of online subjects recruited through Prolific and in-person Brown University students in the Semantic Structure experiment conditions.

Then for circle (C K E), which is three letters, inserting % between each would be C % K % E. But wait, the rectangle's symbols had E % K % C, which is the original square's letters separated by %. So if we apply the same logic to the circle to get the oval, then the oval's symbols would be C % K % E. But wait, let me check if there is another pattern.

Alternatively, maybe the symbols are reversed for the circle. The square is E K C, and the circle is C K E. So the circle's symbols are the reverse of the square's. Then maybe the oval would be the reverse of the rectangle's symbols. The rectangle is E % K % C. Reversing that would be C % K % E. So that would be the oval's symbols. That makes sense because circle is reverse of square, so oval (which is like a stretched circle) would be the reverse of the rectangle's symbols. So rectangle is E % K % C, reverse would be C % K % E for oval.

Yes, that seems plausible. Let me check again. Square to rectangle: inserting % between letters. Circle is reverse of square's letters. So oval would be reverse of rectangle's letters. Therefore, the oval's symbols would be C % K % E. That fits the pattern. So the answer should be C % K % E.

[Output] The answer is: C % K % E"

The very good performance on this example is interesting, but we leave analysis of analogical reasoning in "reasoning models" for future work.

### *A.4. Further details of human performance*

Fig. A.11 shows the difference in performance between online subjects recruited through Prolific and in-person Brown University students in the Semantic Structure experiment conditions. Prolific subjects were paid $1.50 for the task, with Prolific taking an additional $0.50 per subject. This equated to an approximate effective rate of $22 per hour, well above relevant minimum wages. In-person subjects were each paid $10 to reflect the increased time and effort cost. We expect increased performance from the in person subjects for a number of reasons. First, they are more highly remunerated. Second, there may be social pressure to perform well given the presence of a member of the research team. Third, the in-person subjects may not have the decreased attention effects likely experienced by subjects on Prolific who may complete many unrelated and potentially demotivating tasks in a day. Fourth, there is an implicit selection effect on academic performance for students at our university, which is not unrelated to the competencies involved in completing the tasks in the experiment. Indeed, we observe that match to reference answers increases by approximately 0.1–0.2 for the in-person subjects in all but one condition. The exception to this is the Random Finals condition, in which mean match to reference decreases slightly. However, in this condition it is not clear that a decrease in match to reference is objectively worse performance, because in this condition we ask for the drawing corresponding to a final unrelated term, while all previous left-hand terms within the question are related. It is thus not unreasonable to give an answer that differs from what we expect, except insofar as subjects are learning in context from previous questions in the quiz to realize that the final unrelated word should be regarded as irrelevant.

Logistic regression analysis confirms that the in-person subjects outperform the online subjects. This is confirmed with the in-person subjects as the reference class and with the independent variable either being jointly the subject type and quiz class, or the subject type alone (respectively coef −1.1738, P 0.015 and coef −0.6462, P 0.000).

Fig. A.12 below shows the variation in performance among human subjects completing different quizzes. As can be seen, some conditions aggregate over quizzes in which the mean performance is quite stable (for example, Random Finals). Other conditions aggregate over quizzes in which there is a larger variation in performance (for example, Randoms). In the first quiz of the Randoms condition, all respondents score 100%. No particular features of this quiz were identified that would explain this occurrence. However, given that the performance of subsequent quiz-takers is independent, that a significant proportion of all quiz-takers score 100%, and that we have 28 quizzes that each sample a number of respondents, it does not seem unlikely for one quiz to have all perfect scores by chance.

One highly-specific failure mode is present in the human data and deserves special consideration. In only the "*" target domain, participants quite often introduced separator characters into the response (either just ">" or both separator characters, "=>"). This was observed in 11 instances, thus affecting approximately 10% of the responses in that target domain scheme. This issue was not observed in any other target domain scheme. The reason for this is not entirely clear, but could be related to the short length of the target domain elements in this scheme (a right hand term is either 1 or 3 characters in this scheme, which is shorter than the other versions). It is possible that the short length of the right hand terms could lead subjects to perceive the separator characters as being part of the target term, although it is unclear why this would happen even for subjects who successfully ignore the separator characters in three prior responses (which applies to 7 out of 11 such errors).

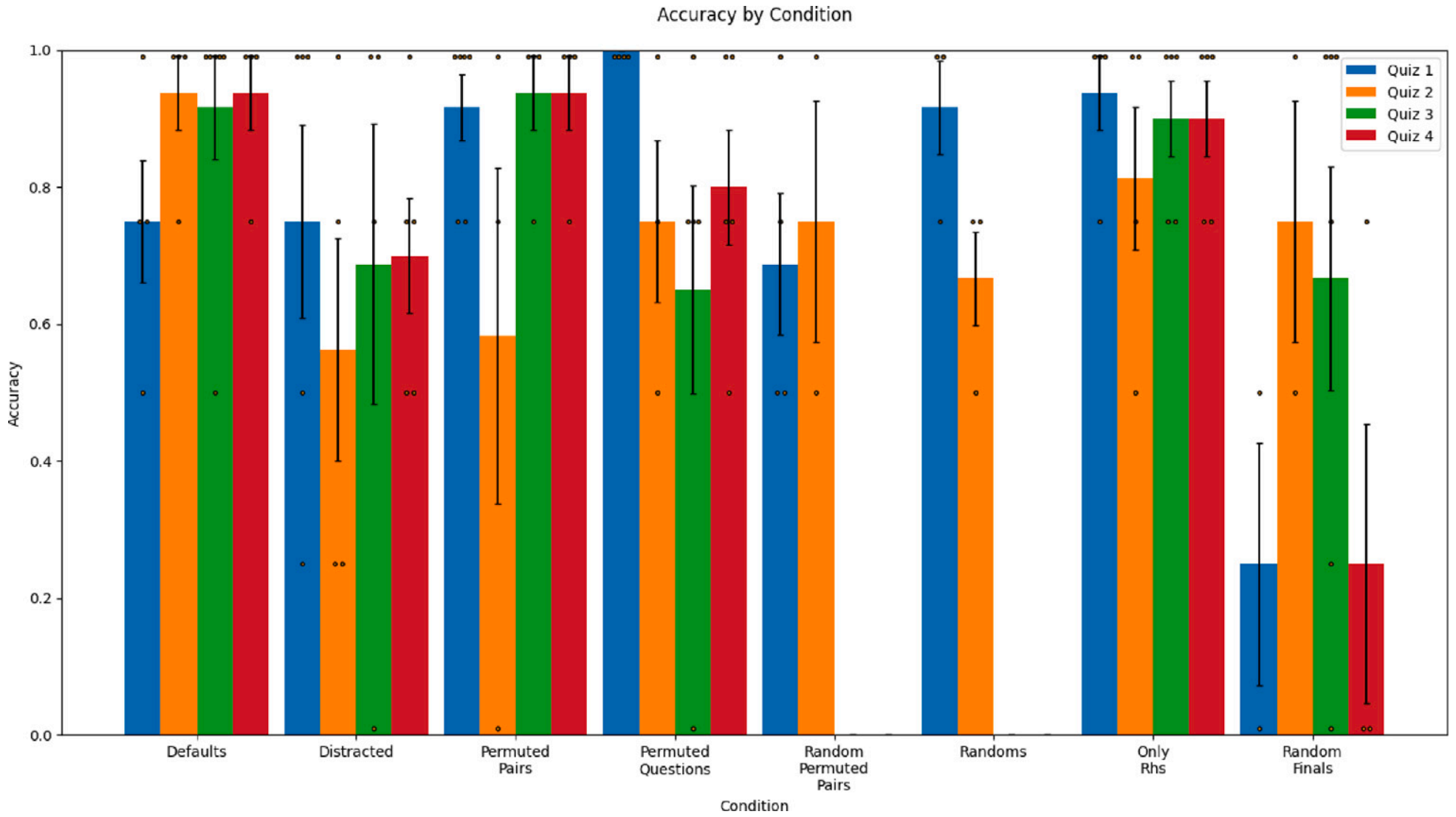

**Fig. A.12.** Human match to reference proportion by quiz across conditions. Error bars show standard errors.

The error rate of the human subjects by target domain type is shown in Fig. A.13. The error rate in the "*", "C K E", and "Q Z I" target domain schemes was essentially equal, with approximately half of this rate in the "c c" target domain scheme. This is not entirely intuitive, but some possible explanations of this can be offered. First, the "c c" right hand terms have the fewest number of distinct characters, thus limiting the option space when answering (there are two distinct characters, compared to 3 or 4 for the other right hand terms). Second, the specific transformations involved (capitalization and adding/removing a letter) are common operations that are encountered more frequently than, say, inserting a special character between existing characters.

As can be seen in Table A.5, about a fifth of the human participants' incorrect responses were simply copies of one of the three right-hand terms presented in the question. Very few participants made a mistake that only reordered the correct answer, whereas about half of all incorrect answers were the wrong combination of characters from the right-hand terms of the task. Note that in all of our target domains, any individual right-hand term only uses characters that are found in at least one of the other three. Thus, the third of incorrect responses from human subjects which did not fall into any of the previous categories included characters which had not been presented in one of the three preceding right-hand terms. This can be explained in some cases by the presence of a distractor that confused a participant into including characters from a right-hand term that used other characters, while in others it can be explained by typos or some other confusion.

In addition to the task questions, subjects were presented with three follow-up questions that asked them to rate their confidence that their answers were correct, describe what they thought the task involved, and describe their strategy for answering the questions.

Subjects employ a mix of strategies in answering the questions. Some subjects explicitly attend to the analogy structure of the left-hand terms. For example, in the Distracted condition, one participant reports that "I tried to find the one that resembled the blank one...ie, red/pink, cat/kitten". By contrast, others focused more on completing the pattern in the right-hand terms. Most subjects robustly ignored the distractor terms in that condition.

Some subjects who report a detailed, correct strategy nevertheless fail to attain a high proportion of answers matching the reference,

**Table A.5**
The distributions of types of errors made by top-performing participants in the Semantic Structure experiment.

| | Copy context | Scrambled | Wrong combination | Other |
|---|---|---|---|---|
| Human | 0.192 | 0.020 | 0.556 | 0.232 |
| GPT-4 | 0.239 | 0.031 | 0.502 | 0.228 |
| Claude 3 Opus | 0.036 | 0.045 | 0.276 | 0.643 |

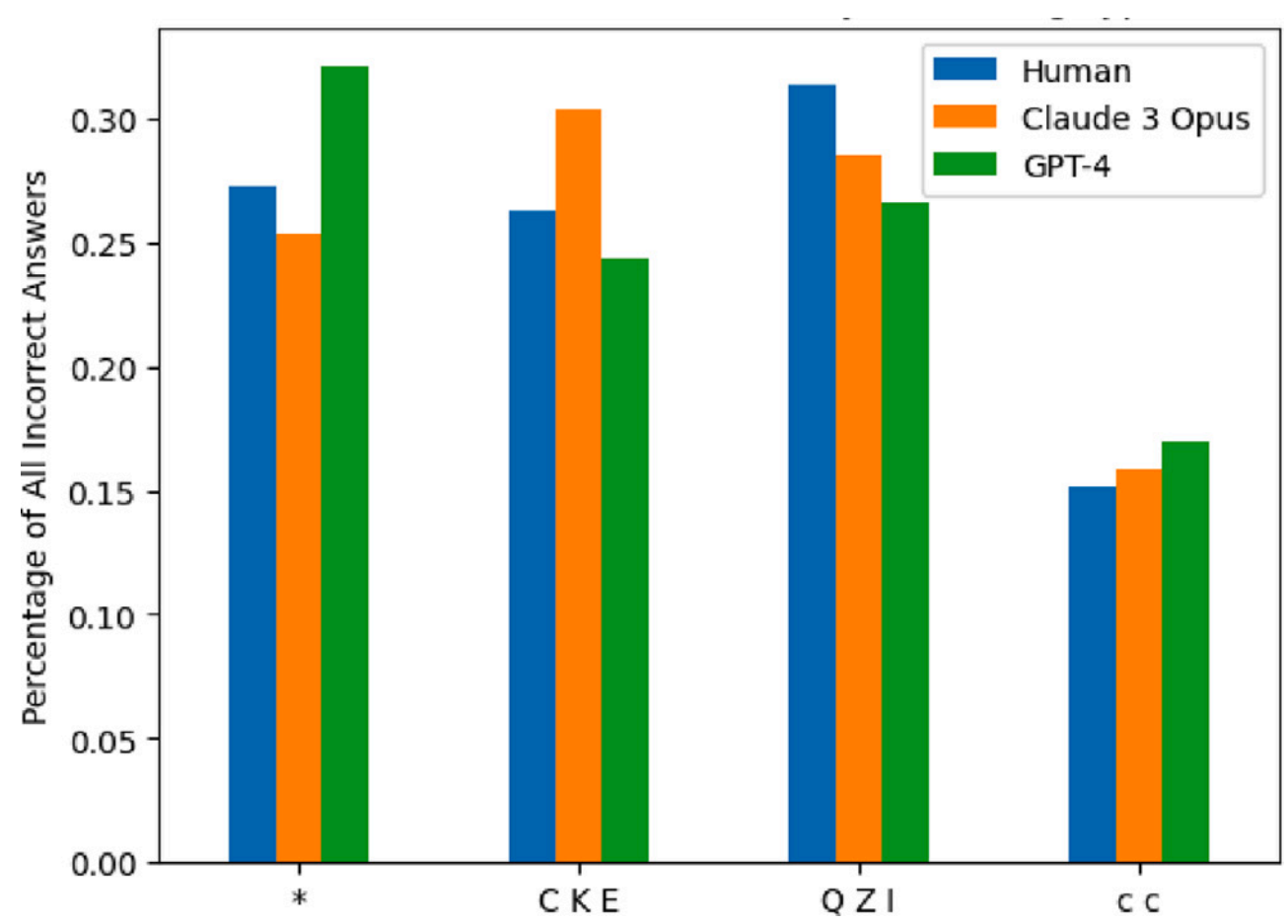

**Fig. A.13.** Percentage of incorrect answers in the Semantic Structure experiment by target domain type.

thus demonstrating that the task is not trivial even for those who are able to fully grasp what it involves. For example, one subject attains a below-average match to reference of 50% in the Distracted condition despite being able to state that the task involves "Looking at other comparable entries to figure out what the answer to the last entry was (dog:puppy::cat:kitten)", and reporting a strategy in which "I tried to find similar pairs of entries and looked at their meanings". By contrast, another subject attains 100% match to reference in the Distracted condition while responding to what the task involved with "I thought it was fun" and reporting a strategy in which "I just compared answers and tried my best to understand what they were and then tried to guess based on my interpretation of the other answers".

In addition to the comparable mean performance, we find similar patterns in the errors made by humans and GPT-4. In Table A.5 and

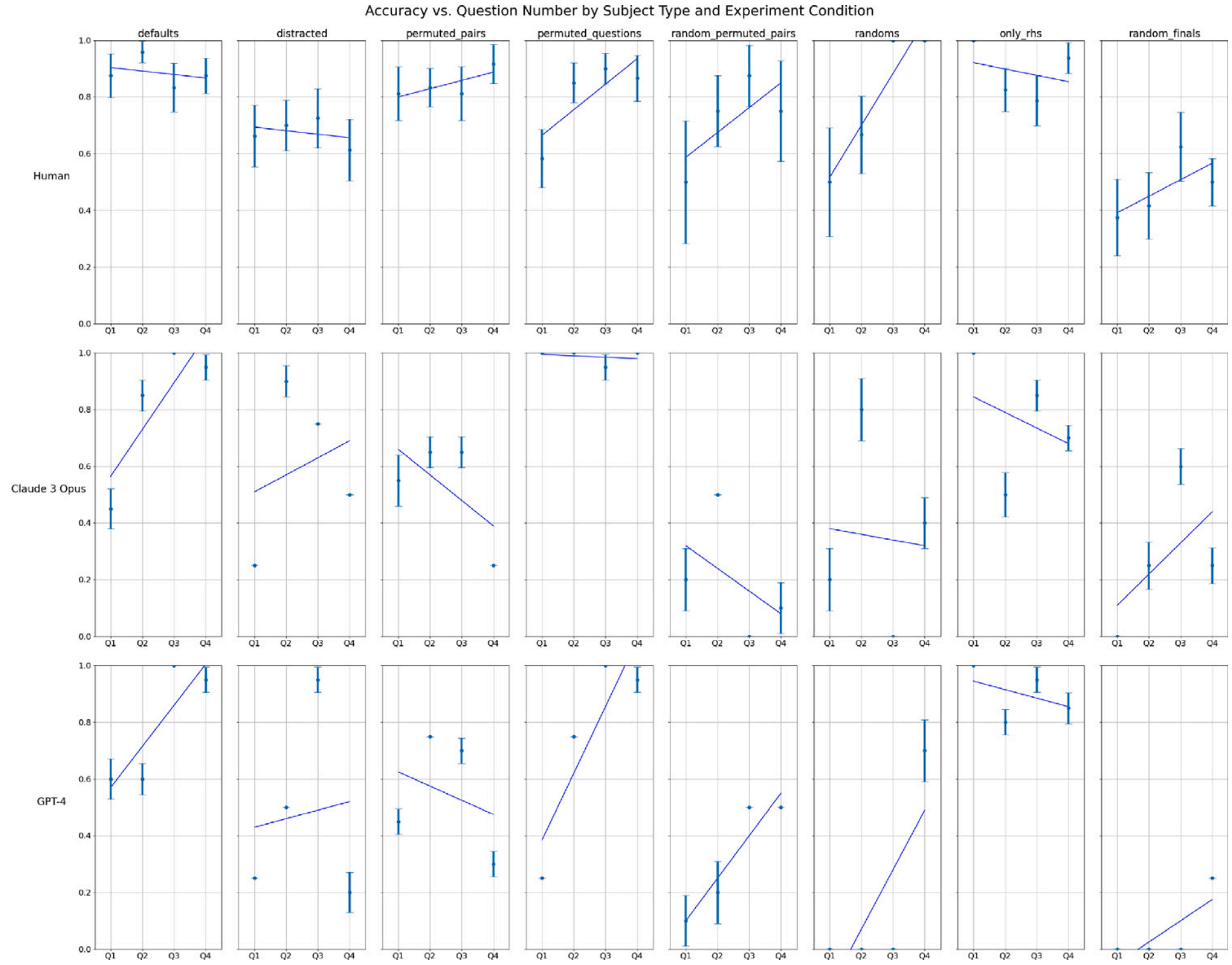

**Fig. A.14.** Improvement in human and LLM proportion match to reference by question number across different conditions. Error bars show standard errors.

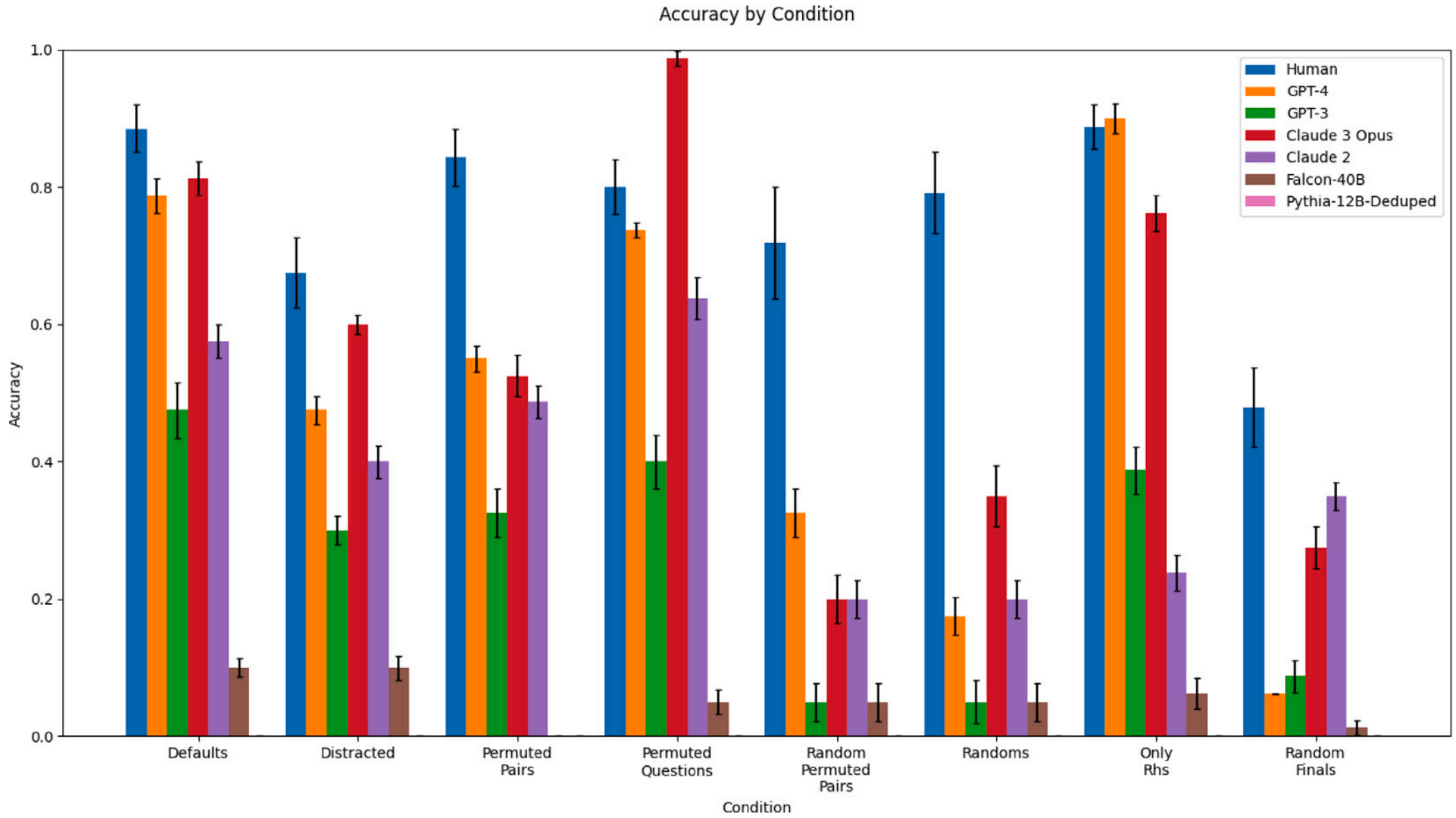

**Fig. A.15.** Performance of all tested models in the Semantic Structure experiment. Error bars show standard errors.

Fig. A.13, one can see that the distribution of errors is comparable both when broken down by target domain type and when broken down by several error classifications we design.

Further, humans and GPT-4 both improve as they see more questions over the course of a quiz. As seen in Fig. A.14, humans display a positive learning trend in 5 out of 8 conditions. GPT-4 displays a positive learning trend in a comparable 6 out of 8 conditions, with one of the conditions in which it does not display improvement resulting from it displaying near-perfect match to reference from start to finish (in the Only RHS condition).

### A.5. Further details of LLM performance

Fig. A.15 shows the performance of all tested models in the Semantic Structure experiment. Some models perform quite poorly, for example Falcon-40B. This model also performs poorly in the Semantic Content experiment, with an illustrative response set shown in Table A.6.

Fig. A.16 shows the variation in the performance of GPT-4 in two conditions (Only RHS and Random Finals) across various small differences in prompting strategy. With small differences in prompting

**Table A.6**
Table showing an illustrative response from Falcon-40B in the CATEGORIAL condition of the Semantic Content experiment. Observe that the model provides incorrect and correct responses to the question, seeming to recognize the form of a correct response but to not reason further about correctness.

| | |
|---|---|
| Prompt | We are conducting an experiment on general reasoning abilities. Below we will show you various words and drawings of each, after which you will need to complete the last drawing. Respond as concisely as possible with only the last drawing.<br>Question 1:<br>chicken => !<br>spider => !<br>cat => *<br>horse => *<br>ant => !<br>dog => *<br>bee => !<br>human => *<br><br>Question 2:<br>car => *<br>tricycle => !<br>motorcycle => *<br>skateboard => !<br>bicycle => !<br>unicycle => |
| Sampled response set | ! [Continuation omitted]<br>! [Continuation omitted]<br>* [Continuation omitted]<br>! [Continuation omitted]<br>* [Continuation omitted] |

**Table A.7**
Table showing an illustrative response from GPT-4 in the NUMERIC condition of the Semantic Content experiment. Observe that the model fails to correctly relate the number of characters to the numerical property of the object, in this case the number of wheels that a car has.

| | |
|---|---|
| Prompt | We are conducting an experiment on general reasoning abilities. Below we will show you various words and drawings of each, after which you will need to complete the last drawing. Respond as concisely as possible with only the last drawing.<br>Question 1:<br>spider => * * * * * * * *<br>human => * *<br>cat => * * * *<br>chicken => * *<br>dog => * * * *<br>horse => * * * *<br>bee => * * * * * *<br><br>Question 2:<br>motorcycle => * *<br>tricycle => * * *<br>bicycle => * *<br>unicycle => *<br>car => |
| Sampled response set | * * *<br>* *<br>* *<br>* *<br>* * |

strategy, performance in the ONLY RHS condition varies between approximately 20% and approximately 100%. Similarly, small differences in the prompting strategy yield performance in the RANDOM FINALS condition that varies between near-zero and close to 40%. Such variation is very significant, but appears in general to be fairly explicable. In the ONLY RHS condition, the majority of the variation appears to come from setting up the prompt in such a way that it is clear that a final term is desired next, as opposed to a new question. In the other conditions, an arrow separator that divides left- and right-hand side terms is the final element of the prompt, thus suggesting that a right-hand term is appropriate as the next token. In the ONLY RHS condition, this trailing separator was initially not present, and thus the models often responded by beginning a new question rather than by completing the last question presented. Re-introducing arrow separators and making other small changes designed to more clearly indicate when a question has not yet been completed eliminates these kinds of errors and drastically increases performance. In the RANDOM FINALS condition, a significant improvement comes from changing the instruction sentence from one that specifies that a drawing of the left-hand side is requested, to an instruction sentence specifying that various patterns will be shown after which the last should be completed. This is reasonable, as in this condition the final left-hand term is misleading and so an instruction focusing attention on it is expected to reduce performance. As expected, no performance improvement is observed when replacing the set of random final words with different ones. Finally, a performance boost is observed when adding additional newlines (instead of only having a clear line between each question, we now also include a clear line between each line of a question). It is not clear why this should improve performance.

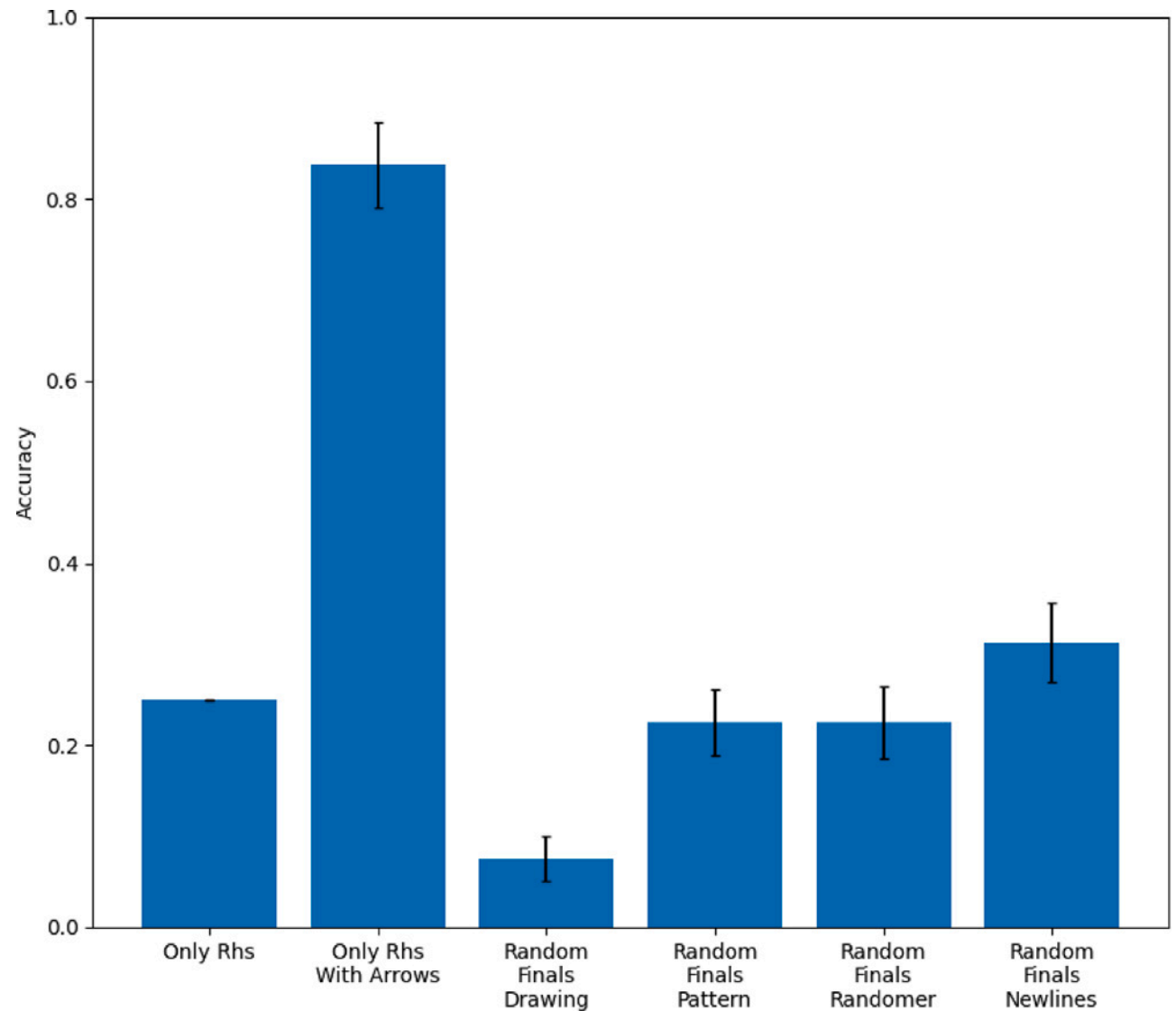

**Fig. A.16.** Dependence of GPT-4 match to reference proportion on prompt variations. Error bars show standard errors.

**Table A.8**
Table showing a sanity check that GPT-4 fails to reason about the number of characters in the expected way. Settings: temperature 1, maximum length 256, top P 1.

| | |
|---|---|
| Prompt | * * * + * = * * * *<br>* * * * * _ * * = * * *<br>* * * * * * _ * * * = |
| Expected result | * * * |
| Responses | * * * * * *<br>* * * * * *<br>* * * * * * * |

## Data availability

The code and data used for generating all experiments are available at https://github.com/smusker/LLM_Analogical_Reasoning.